\documentclass[screen]{acmart} 

\AtBeginDocument{%
  }

\setcopyright{none}  
\copyrightyear{2025}
\acmYear{2025}
\acmDOI{XXXXXXX.XXXXXXX}

\usepackage{enumitem} 
\usepackage{breakurl}

\begin{document}

\title{Deep Learning Approaches for Multimodal Intent Recognition: A Survey}


\author{Jingwei Zhao}
\authornote{Both authors contributed equally to this research.}
\affiliation{%
  \institution{Beijing University of Posts and Telecommunications}
  \city{Beijing}
  \country{China}}
\email{zhaojingwei@bupt.edu.cn}
\author{Yuhua Wen}
\authornotemark[1]
\affiliation{%
  \institution{Beijing University of Posts and Telecommunications}
  \city{Beijing}
  \country{China}}
\email{yuhuawen@bupt.edu.cn}

\author{Qifei Li}
\affiliation{%
  \institution{Beijing University of Posts and Telecommunications}
  \city{Beijing}
  \country{China}}
\email{liqifei@bupt.edu.cn}

\author{Minchi Hu}
\affiliation{%
  \institution{Beijing University of Posts and Telecommunications}
  \city{Beijing}
  \country{China}}
\email{minchihu@bupt.edu.cn}

\author{Yingying Zhou}
\affiliation{%
  \institution{Beijing University of Posts and Telecommunications}
  \city{Beijing}
  \country{China}}
\email{yingyingzhou@bupt.edu.cn}

\author{Jingyao Xue}
\affiliation{%
  \institution{Beijing University of Posts and Telecommunications}
  \city{Beijing}
  \country{China}}
\email{wuwuxjy@bupt.edu.cn}

\author{Junyang Wu}
\affiliation{%
  \institution{Beijing University of Posts and Telecommunications}
  \city{Beijing}
  \country{China}}
\email{wujunyang128@gmail.com}

\author{Yingming Gao}
\affiliation{%
  \institution{Beijing University of Posts and Telecommunications}
  \city{Beijing}
  \country{China}}
\email{yingming.gao@bupt.edu.cn}

\author{Zhengqi Wen}
\affiliation{%
  \institution{Tsinghua University}
  \city{Beijing}
  \country{China}}
\email{zqwen@tsinghua.edu.cn}

\author{Jianhua Tao}
\affiliation{%
  \institution{Tsinghua University}
  \city{Beijing}
  \country{China}}
\email{jhtao@tsinghua.edu.cn}

\author{Ya Li}
\authornote{Corresponding author.}
\affiliation{%
  \institution{Beijing University of Posts and Telecommunications}
  \city{Beijing}
  \country{China}}
\email{yli01@bupt.edu.cn}
\thanks{This work is supported by the National Key R\&D Program of China under Grant No.2024YFB2808802.}

\renewcommand{\shortauthors}{J. Zhao et al.}

\begin{abstract}

Intent recognition aims to identify users' underlying intentions, traditionally focusing on text in natural language processing. With growing demands for natural human-computer interaction, the field has evolved through deep learning and multimodal approaches, incorporating data from audio, vision, and physiological signals. Recently, the introduction of Transformer-based models has led to notable breakthroughs in this domain. This article surveys deep learning methods for intent recognition, covering the shift from unimodal to multimodal techniques, relevant datasets, methodologies, applications, and current challenges. It provides researchers with insights into the latest developments in multimodal intent recognition (MIR) and directions for future research.
\end{abstract}




\maketitle

\section{Introduction}

Intent recognition is a computational process that aims to infer a user's underlying goal or objective from their textual, spoken, or other interaction data~\cite{atuhurra2024domain}. With the rapid development of AI technology, especially the emergence of deep learning and large-scale corpora, intent recognition has rapidly developed and shown great potential and value in many application areas, such as human-computer interaction~\cite{jain2019probabilistic,xu2023intent}, dialog systems~\cite{schuurmans2019intent,liu2019review}, healthcare~\cite{zhang2023conco,wang2022recognizing}, and recommendation systems~\cite{li2021intention,jannach2024survey,chen2022intent}. In conversational AI, such as virtual assistants like Siri or Alexa, accurate intent recognition enables personalized and context-aware responses, significantly enhancing the user experience. In smart homes, it helps customize automation based on user preferences~\cite{rafferty2017activity}. Similarly, in automotive systems, intent recognition supports safer, more intuitive driver-assistance interactions~\cite{berndt2008continuous}. The transformative potential of intent recognition lies in its ability to bridge the gap between human intent and machine behavior, driving innovation and improving efficiency across these domains. As society becomes increasingly reliant on intelligent systems, the demand for powerful, accurate, and adaptable intent recognition technologies continues to grow, prompting researchers to explore new methods to address these challenges.

In the early stages of intent recognition research, researchers focused on unimodal intent recognition represented by text modality~\cite{allen1980analyzing}, which relies solely on information from a single modality to understand user intent. Early research relied on rule matching and manual feature engineering, such as template-based text parsing or keyword extraction after speech-to-text conversion~\cite{holtgraves2008automatic}, but these methods have insufficient generalization ability in the face of complex semantics and diverse expressions. With the development of deep learning technology, unimodal intent recognition has also made significant progress. Pre-trained models such as BERT have substantially enhanced the representation and understanding of semantic content in user inputs~\cite{chen2019bert, wu2021label}, boosting performance across numerous tasks~\cite{huggins2021practical, pearce2023build}. LLM-based methods can effectively address intent recognition challenges in zero-shot and few-shot learning. Despite significant advancements in intent recognition technology, current mainstream unimodal intent recognition methods still face numerous challenges. Single-modal data, such as independent audio, text, or visual information, often fails to provide sufficient information for accurately and robustly identifying complex user intentions. For example, in the audio modality, factors such as background noise, accent differences, and variations in speaking speed can severely impair audio intent recognition accuracy. In the text modality, synonyms, polysemous words, and ambiguity caused by missing context often make it difficult for systems to understand users' true intentions. In the visual modality, issues such as partial occlusion, lighting changes, and perspective limitations may also result in insufficient visual information, thereby affecting the reliability of intent recognition. These inherent limitations mean that single-modality intent recognition suffers significantly in terms of accuracy and robustness when faced with complex, uncertain, or information-deficient scenarios, making it difficult to meet the demands of practical applications.

To overcome the limitations of single-modal intent recognition, multimodal intent recognition has emerged, aiming to fuse information from different modalities to enhance the performance of intent recognition. Modalities of multimodal intent recognition typically involve, but are not limited to, audio, text, visual data (e.g., gestures, facial expressions, or eye gaze), and physiological signals \cite{zhao2024learning,SLANZI2017eyeEEGMIR}. In recent years, the widespread adoption of sensor technology and data collection devices, such as smartphones, cameras, and wearable devices, has made multimodal data collection more convenient, resulting in a significant increase in the volume and diversity of data, which provides a solid foundation for research. Researchers have advanced the development of multimodal intent recognition research by leveraging the complementary advantages of multimodal data, significantly improving the accuracy of human intent recognition in real-world scenarios. Through the use of deep learning techniques such as modality alignment and cross-modal fusion \cite{TCL-MAP}, models have enhanced their ability to integrate and understand multimodal information, providing new technical pathways for the evolution of human-computer interaction toward more natural and intelligent directions. However, multimodal intent recognition presents new challenges. How to effectively fuse data from different modalities to achieve complementary advantages between modalities; how to handle the heterogeneity, asynchrony, and conflicting information across modalities; how to construct models that can capture the complex associations and intrinsic logic of multimodal data while addressing missing or incomplete modalities—these are the key issues that the field of multimodal intent recognition urgently needs to address. These challenges provide a broad research landscape and abundant exploration directions for this field.

\begin{figure}[htbp]
  \centering
  \includegraphics[width=1\linewidth]{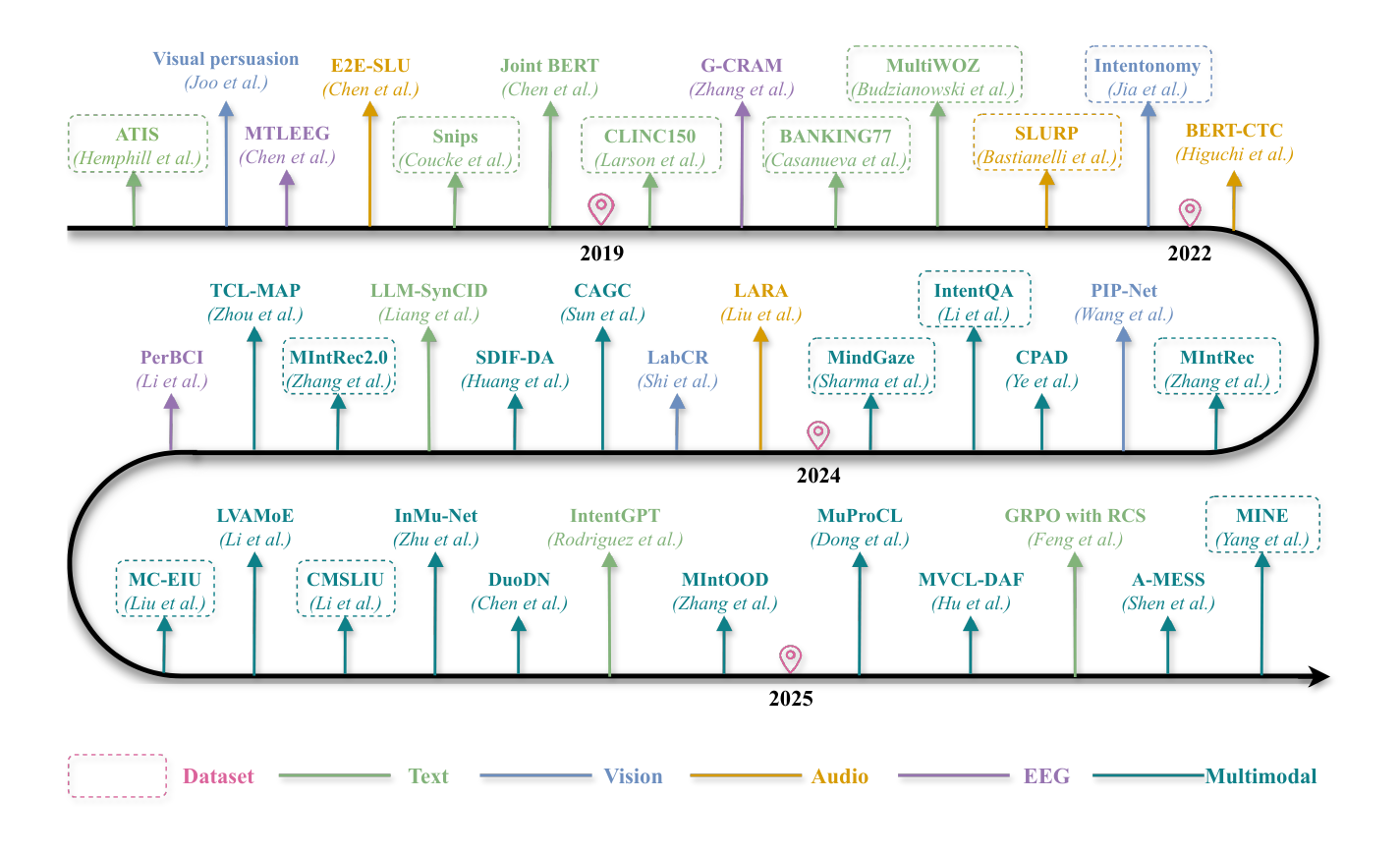}
  \caption{Representative Methods and Datasets for Intent Recognition.}
  \label{fig:timeline}
\end{figure}

\begin{figure}[htbp]
  \centering
  \includegraphics[width=1\linewidth]{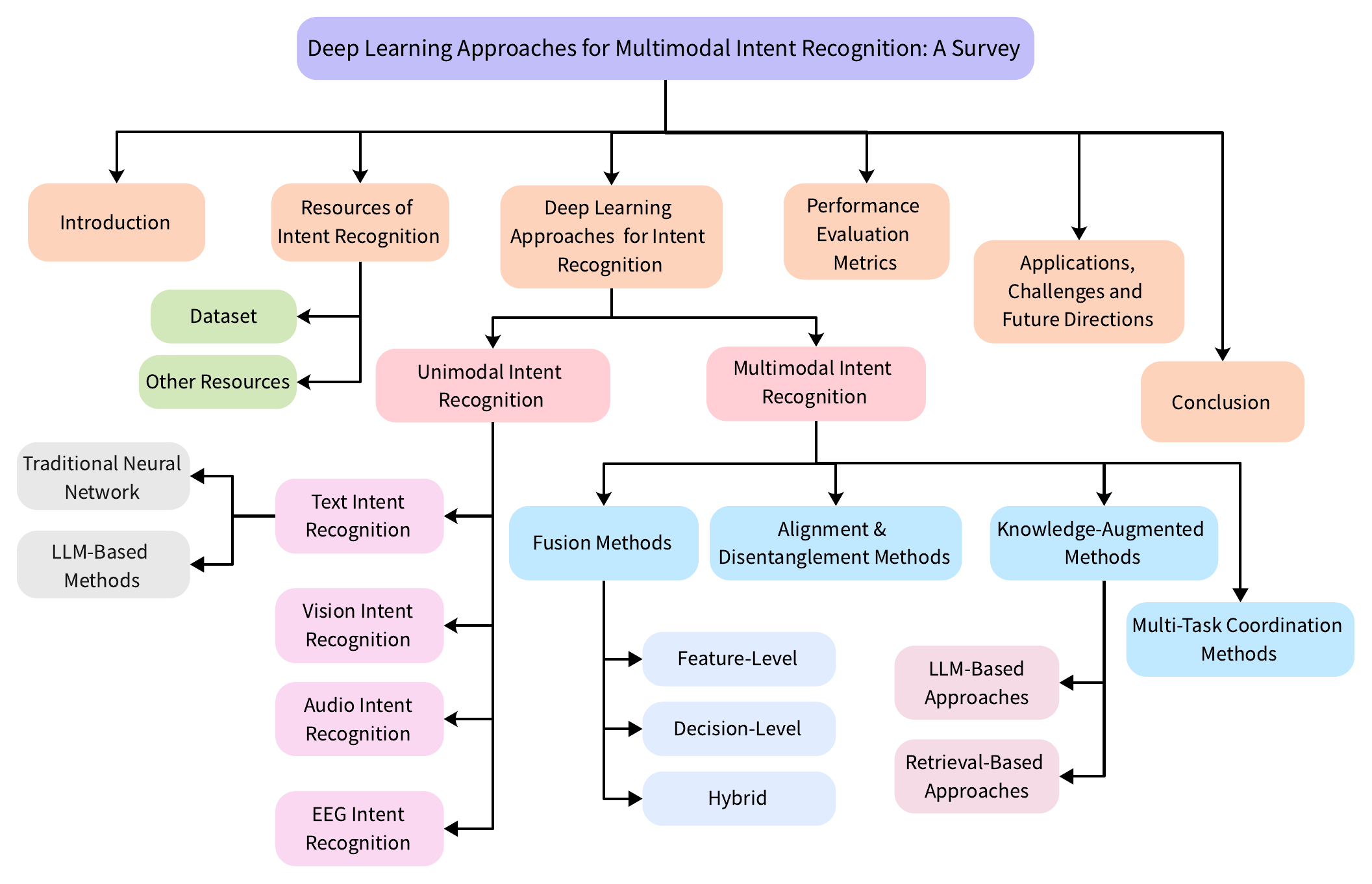} 
  \caption{Organizational Structure of the Survey.}
  \label{fig:organize}
\end{figure}

The primary goal of this paper is to provide a comprehensive summary and in-depth analysis of the evolution of intent recognition techniques from unimodal to multimodal, aiming to systematically sort out the technological development, application scenarios, and challenges in this field. The article first combs through the widely used intent recognition datasets, covering unimodal and multimodal. It then reviews common unimodal intent recognition methods, including text, vision, audio, and EEG modalities. For example, textual intent recognition relies on natural language processing techniques, audio intent recognition employs speech signal processing, while visual and EEG intent recognition utilize image processing and neural signal analysis to capture user intent. Then, focusing on the rapid rise of multimodal intent recognition in recent years, the article analyzes and discusses the approaches of multimodal modeling strategies. In Figure \ref{fig:timeline}, we present representative works for intent recognition. Meanwhile, the paper summarizes the evaluation metrics in the field of multimodal intent recognition, which provides a standardized reference framework for researchers. In terms of applications, multimodal intent recognition has shown many prospects in human-computer interaction, education, automotive systems, and healthcare~\cite{xu2021noninvasive}. In addition, multiple challenges still facing current research and new research trends are also included to provide useful references for researchers and developers. The organizational structure of this survey is shown in Figure \ref{fig:organize}.

The significant contributions of this survey are outlined as follows:

\begin{itemize}
\item We present the first systematic review that traces the development of intent recognition from early unimodal approaches to modern multimodal techniques, offering a structured and comparative perspective on the evolution of this field.
\item We collate and analyze benchmark datasets and evaluation metrics, covering both unimodal and multimodal settings, to offer researchers a standardized foundation for experimentation and comparison.
\item We summarize the application scenarios (e.g., HCI, healthcare, automotive systems) and highlight key challenges such as modality heterogeneity, synchronization, and robustness to missing data. Based on these, we discuss emerging trends and future research directions.
\end{itemize}

\section{Resources of Intent Recognition Research}

\subsection{Dataset}
Datasets in the field of intent recognition provide rich samples and annotated resources for research. They serve as fundamental assets for advancing model training, algorithm validation, theoretical verification, and pattern discovery, playing a pivotal role in driving progress within the domain of intent understanding. They provide annotated resources for diverse scenarios, but their intent categories often lack unified standards. Inspired by the classification approach of the MIntRec dataset\cite{mintrec} and the distribution patterns of intentions across domains, we categorize coarse-grained labels into three primary classes: Emotion and Attitude, Goal Achievement, and Information and Declaration. This synthesized taxonomy, illustrated in Table \ref{tab:intent_classes}, abstracts away domain-specific variations to reveal foundational intent types.

\begin{table}[htbp]
  \centering
  \caption{Summary of Common Coarse-Grained and Fine-Grained Intent Categories.}
  \label{tab:intent_classes}
  \begin{tabular}{l p{0.34\textwidth} p{0.34\textwidth}}
    \toprule
    \textbf{Coarse-Grained Intent} & \textbf{Definition} & \textbf{Fine-Grained Intent} \\
    \midrule
    Emotion and Attitude & Express emotional states, subjective feelings or attitudes towards things. & Complain, Praise, Thank, Agree, Oppose, Doubt, Refuse, Warn, Antipathy, etc.~\cite{mintrec,mintrec2,MC-EIU,LI2024MBCFNET} \\ 
    
    Goal achievement & Clearly express the hope to achieve a certain goal, request the other party to perform an operation or provide a service. & Ask for help, Advise, Suggest, Invite, Plan, Arrange, Inform, Request, Select, Prevent, BookRestaurant, PlayMusic, AddToPlaylist, Query, Directive, etc.~\cite{mintrec,mintrec2,Multiwoz,MC-EIU,LI2024MBCFNET,Snips} \\ 
    
    Information and Declaration & State facts or describe situations objectively. & Virtues, Family, Health, Power, Openness to experience, etc.~\cite{intentonomy} \\
    \bottomrule
  \end{tabular}
\end{table}

Table \ref{tab:dataset} systematically compares these datasets across key dimensions: sample size, number of intent classes, and modality. This quantitative overview reveals disparities in data scale and diversity.

\begin{table}[htbp]
  \centering
    \caption{Summary of Commonly Used Intent Recognition Dataset}
    \label{tab:dataset}
    \begin{tabular}{c c c c c c c p{3.32cm}}
    \toprule
    \textbf{Dataset} & \textbf{Year} & \textbf{Samples} & \textbf{Nums Int.} & \textbf{Modal} & \textbf{Language} & \textbf{Avail.} & \textbf{URL} \\
    \midrule
    ATIS\cite{ATIS} & 1990 & 5,871 & 17 & T & EN & Public & \url{https://datasets.activeloop.ai/docs/ml/datasets/atis-dataset/} \\ 
    Snips\cite{Snips} & 2018 & 14,484 & 7 & T & EN & Public & \url{https://github.com/sonos/nlu-benchmark} \\
    CLINC150\cite{CLINC} & 2019 & 23,700 & 150 & T & EN & Public & \url{https://github.com/clinc/oos-eval} \\
    MultiWOZ\cite{Multiwoz} & 2020 & 10,438 & 13 & T & EN & Public & \url{https://github.com/budzianowski/multiwoz} \\
    BANKING77\cite{BANKING77} & 2020 & 13,083 & 77 & T & EN & Public & \url{https://github.com/PolyAI-LDN/task-specific-datasets7} \\
    SLURP\cite{bastianelli2020slurp} & 2020 & 72,277 & 46 & A & EN & Public & \url{https://github.com/pswietojanski/slurp} \\
    Intentonomy\cite{intentonomy} & 2021 & 14,455 & 28 & V & EN & Public & \url{https://github.com/kmnp/intentonomy} \\
    MIntRec\cite{mintrec} & 2022 & 2,224 & 20 & VAT & EN & Public & \url{https://github.com/thuiar/MIntRec} \\
    IntentQA\cite{IntentQA} & 2023 & 16,297 & / & VT & EN & Public & \url{https://github.com/JoseponLee/IntentQA} \\
    MindGaze\cite{MindGaze} & 2023 & 3,600 & 2 & EEG, EM & / & Limited & \url{https://zenodo.org/records/8239062} \\
    MIntRec2.0\cite{mintrec2} & 2024 & 15,040 & 30 & VAT & EN & Public & \url{https://github.com/thuiar/MIntRec2.0} \\
    MC-EIU\cite{MC-EIU} & 2024 & 56,012 & 9 & VAT & EN, ZH & Public & \url{https://huggingface.co/datasets/YulangZhuo/MC-EIU} \\
    CMSLIU\cite{CMSLIU} & 2024 & 5,520 & 6 & AT, EEG & ZH & Limited & \url{https://drive.google.com/drive/folders/1w76HxNj4zWK3snpdjlr9-aDNRddOIrlD} \\
    MINE\cite{yang2025uncertain} & 2025 & 20,168 & 20 & VAT & EN & Public & \url{https://github.com/yan9qu/CVPR25-MINE} \\
    \bottomrule
    \end{tabular}

    \raggedright
    EN: English, ZH: Chinese, T: Text, V: Vision, A: Audio, EM: Eye Movement
\end{table}

\begin{enumerate}[label=(\arabic*)] 

    \item \textbf{Text datasets}: As one of the earliest widely explored modalities in intent recognition research, text modality, relying on the rich semantic information carried in natural language, provides an important foundation for modeling and classifying users' intent. Several high-quality text datasets have been widely used for this task. The ATIS~\cite{ATIS} dataset is comprised of 5,871 English texts concerning flight-related queries, which are subdivided into 17 distinct categories of intent. The dataset is primarily utilized for the evaluation of intent recognition capabilities within closed domains. The SNIPS~\cite{Snips} is a crowdsourced dataset containing 14,484 user queries categorized into 7 intents with varying complexity levels. The CLINC150~\cite{CLINC} dataset collects 23,700 query statements, of which 22,500 cover 150 valid intent categories and are categorized into 10 generic domains, and 1,200 cross-domain samples, which are suitable for evaluating open-domain intent recognition systems. MultiWOZ~\cite{Multiwoz} is a multi-domain, multi-round dialogue corpus containing 10,438 samples (8,438 conversations) with an average of 14 turns per conversation, covering seven service domains including restaurants and hotels. Its evolution includes MultiWOZ 2.1~\cite{eric2019multiwoz} for annotation corrections and MultiWOZ 2.2~\cite{zang2020multiwoz} fixing 17.3\% of dialogue state annotations while standardizing slot representations. The BANKING77~\cite{BANKING77} dataset focuses on the banking service domain and contains 13,083 customer service requests, which are subdivided into 77 types of intentions, emphasizing the fine-grained intent recognition under a single domain.

    \item \textbf{SLURP}~\cite{bastianelli2020slurp}: SLURP(Spoken Language Understanding Resource Package) is a comprehensive dataset for Spoken Language Understanding (SLU) that contains approximately 72,277 audio recordings of single-turn user interactions with a home assistant. Each audio recording is annotated with three levels of semantics: Scenario, Action, and Entities. The dataset spans 18 different scenarios, with 46 defined actions and 55 distinct entity types. The dataset is divided into 50,628 training audio files, 8,690 development audio files, and 13,078 test audio files. SLURP is designed to be substantially larger and more linguistically diverse than existing SLU datasets, making it a challenging benchmark for developing and evaluating SLU systems.
    
    \item \textbf{Intentonomy} ~\cite{intentonomy}: Jia et al. proposed a dataset called Intentonomy, which contains 14,455 high-quality images, each of which is annotated with one or more of 28 human-perceived intentions. These intent labels are based on a systematic social psychology taxonomy proposed by psychology experts, covering 9 supercategories such as "virtue", "self-actualization", "openness to experience", etc. The dataset contains 12,740 training images, 498 validation images, and 1217 test images.

    \item \textbf{MIntRec}~\cite{mintrec}: MIntRec is a new dataset created for multimodal intent recognition tasks, containing 2,224 high-quality samples, each of which includes text, video, and audio modalities, and has multimodal annotations for 20 intent categories. The dataset is built based on data collected from the TV series Superstore, and coarse-grained and fine-grained intent taxonomies are designed. The coarse-grained taxonomy includes "expressing emotions or attitudes" and "achieving goals". The MIntRec dataset randomly shuffles the video clips and divides them into training set, validation set and test set in a ratio of 3:1:1. The training set contains 1,334 samples, and the validation set and test set contain 445 samples respectively.

    \item \textbf{IntentQA}~\cite{IntentQA}: IntentQA is a large-scale VideoQA dataset specifically designed for intent reasoning tasks. It is constructed based on the NExT-QA dataset, focusing on inference-oriented QA types such as Causal Why, Causal How, Temporal Previous, and Temporal Next. The dataset contains 4,303 videos and 16,297 question-answer pairs, covering a diverse range of daily social activities. The key feature of IntentQA is its emphasis on contextual reasoning, where the same action may imply different intent depending on the situational context. The dataset is carefully annotated to ensure high quality, with each sample validated by multiple annotators.

    \item \textbf{MindGaze}~\cite{MindGaze}: MindGaze is the first multimodal dataset for navigation intent (free viewing) and information intent (goal-directed search) prediction, combining EEG to measure brain activity and eye-tracking to record gaze patterns. It features 120 diverse industrial scenes (e.g., assembly units, labs, garages) with five target tools, designed to simulate realistic visual search tasks.
    
    \item \textbf{MIntRec2.0}~\cite{mintrec2}: MIntRec2.0 is a large-scale, multimodal, multi-party dialogue intent recognition benchmark dataset, containing 1,450 high-quality dialogues with a total of 15,040 samples. Each sample is labeled with 30 fine-grained intent categories, covering text, video, and audio modalities. In addition to over 9,300 samples within the range, it also includes over 5,700 out-of-range samples that appear in multi-round dialogue scenarios. These samples naturally appear in the open scenarios of the real world, enhancing their practical applicability. Furthermore, it also provides comprehensive information about the speaker in each discourse, enriching its utility in multi-party dialogue research. The dataset is partitioned into training, validation, and testing sets, maintaining an approximate ratio of 7:1:2 for both dialogues and utterances.
    
    \item \textbf{MC-EIU}~\cite{MC-EIU}: MC-EIU is a dataset for multimodal dialogue emotion and intention joint understanding tasks. It contains two languages, English and Mandarin, and covers three modalities of information: text, audio, and video. It contains seven emotion categories and nine intent categories with a total of 56,012 utterances from three English TV series and four Mandarin TV series. The MC-EIU dataset is divided into the training set, the validation set, and the test set in a ratio of 7:1:2.
    
    \item \textbf{CMSLIU}~\cite{CMSLIU}: The CMSLIU dataset is a multimodal dataset specifically designed for Chinese intent recognition. It combines physiological signals (EEG) and non-physiological signals (audio and text) to study how humans use the same text to express different intentions in conversations. The dataset contains 184 sentences of text and a total of 15 video clips. Each clip is followed by a varying number of text reading tasks. This dataset can be used to study the influence of emotions on intent recognition and the role of electroencephalogram (EEG) information in intent recognition.

    \item \textbf{MINE}~\cite{yang2025uncertain}: MINE (Multimodal Intention and Emotion Understanding in the Wild) is a large-scale dataset focused on multimodal intention and emotion understanding in social media contexts. Collected from real-world social media platform Twitter, it comprises over 20,000 social media posts encompassing text, images, videos, and audio, annotated with 20 intention categories and 11 emotion labels. The dataset uniquely captures real-world modality incompleteness (e.g., posts containing only text and images) and implicit intention-emotion correlations (e.g., the "Comfort" intention often co-occurs with "Sad" emotion). MINE provides the first real-world benchmark for studying multimodal fusion, missing modality handling, and joint emotion-intention modeling.
    
\end{enumerate}

\subsection{Other Resources}
In the field of intent recognition, academia and industry have developed multiple powerful open-source tools and platforms that provide researchers with comprehensive support ranging from theoretical verification to practical application. These systems not only cover traditional closed intent recognition tasks but also offer innovative solutions to the problem of discovering unknown intent in open-world scenarios.

\begin{enumerate}[label=(\arabic*)] 
    \item \textbf{TEXTOIR}~\cite{textoir} is an integrated visualization platform for textual open intent recognition. Its main functions include open intent detection and open intent discovery. The system supports a variety of advanced algorithms and benchmark datasets, and builds a unified pipeline framework, thereby enabling fully automatic processing from known intent recognition to unknown intent clustering. The platform provides standardized interfaces and visualization tools to facilitate model training, evaluation, and result analysis. This significantly enhances the implementation and research of open intent recognition tasks in real-world dialogue systems. More detailed information can be found on \url{https://github.com/thuiar/TEXTOIR}.

    \item \textbf{Rhino}, developed by Picovoice, is an open-source speech-to-intent engine leveraging deep learning to infer user intents and slots from voice commands in real time on-device, eliminating the need for cloud connectivity. Tailored for domain-specific voice interactions, it is ideal for applications in IoT, smart homes, and embedded systems. Utilizing context files to define commands, Rhino supports multiple languages and cross-platform deployment (e.g., Linux, Android, iOS, Web). Its lightweight design ensures efficient processing with minimal resource demands, prioritizing user privacy. Integrated with the Picovoice Console for customizable context modeling, Rhino is a robust tool for intent recognition in voice-driven, resource-constrained environments (\url{ https://github.com/Picovoice/rhino}).

\end{enumerate}

\section{Unimodal Intent Recognition}
The core challenge of unimodal intent recognition lies in extracting intent-related features from heterogeneous data. Text modality relies on language models to parse semantic structures, while visual modality requires modeling spatial and temporal relationships. Audio modality captures paralinguistic information through acoustic analysis, and EEG necessitates decoding spatiotemporal patterns in neural signals. The distinct data characteristics and technical requirements of each modality form the foundation for diverse intent recognition approaches. This section systematically investigates intent recognition methodologies across four distinct modalities: textual, visual, auditory, and EEG. A summary of the representative approaches of unimodal intent recognition is shown in Table \ref{tab:MIR-single}.

\begin{table}[htbp]
  \centering
  \caption{Summary of Representative Approaches of Unimodal Intent Recognition}
  \label{tab:MIR-single}
    \begin{tabular}{p{2.8cm} p{1.8cm} c p{5.3cm} c}
    \toprule
    \textbf{References} & \textbf{Dataset} & \textbf{Modality} & \multicolumn{1}{c}{\textbf{Method}} & \textbf{Performance} \\
    \midrule
    Xu et al.~\cite{xu2013convolutional} & Comm. & T & CNN-based joint intent detection and slot filling. & 93.6\% Accuracy \\
    Hakkani et al.~\cite{hakkani2016multi} & Comm. & T & Multi-domain sequence tagging for intent detection and slot filling. & 95.4\% Accuracy \\
    Yao et al.~\cite{yao2013recurrent} & ATIS & T & RNN language understanding with future words and named entities. & 96.6\% F1-score \\
    Goo et al.~\cite{goo2018slot} & Snips & T & Slot-gated attention for joint slot filling and intent detection. & 97.0\% Accuracy \\
    Chen et al.~\cite{chen2019bert} & Snips & T & BERT-based joint intent classification and slot filling. & 98.6\% Accuracy \\
    Rodriguez et al.~\cite{rodriguez2024intentgpt} & BANKING & T & Training-free intent discovery using LLMs. & 77.21\% Accuracy \\
    Liang et al.~\cite{liang2024synergizing} & BANKING & T & LLM-SLM synergy via contrastive space alignment. & 86.80\% Accuracy \\
    Park et al.~\cite{park2024dynamic} & BANKING & T & Dynamic intent tag refinement with LLM semantic optimization. & 87.30\% Accuracy \\
    Feng et al.~\cite{feng2025improving} & MultiWOZ & T & RL-based intent detection with GRPO and chain of thought. & 93.25\% Accuracy \\
    Jia et al.~\cite{jia2021intentonomy} & Intentonomy & V & Object-centric image decomposition with hashtag incorporation. & 28.88\% Ave F1 \\
    Wang et al.~\cite{wang2023prototype} & Intentonomy & V & Visual intent recognition via prototype learning. & 41.85\% Ave F1 \\
    Shi et al.~\cite{shi2023learnable} & Intentonomy & V & Hierarchical transformer for intent category relationships. & 35.93\% mAP \\
    Chen et al.~\cite{chen2018spoken} & Customer care dataset & A & End-to-end speech-to-intent mapping. & 98.07\% Accuracy \\
    Wang et al.~\cite{wang2021fine} & SLURP & A & Audio encoder fine-tuning beyond ASR tasks. & 89.38\% Accuracy \\
    Ray et al.~\cite{ray2021listen} & SLU and ASR & A & RNN-T system with audio-to-intent embeddings. & 87.70\% Accuracy \\
    Higuchi et al.~\cite{higuchi2022bert} & SLURP & A & BERT adaptation for connectionist temporal classification. & 87.80\% Accuracy \\
    \bottomrule
  \end{tabular}
\end{table}

\subsection{Text Intent Recognition}
Intent recognition constitutes a fundamental component of Natural Language Understanding (NLU), which identifies the expressed intent categories inherent within user utterances. In existing research, text modality is the main input form and focus of research in the field of intent recognition due to its clear structure and high information density. This section will follow the development of technology, illustrating the technical evolution of text intent recognition from traditional machine learning methods to deep learning methods.

\begin{figure}[htbp]
  \centering
  \includegraphics[width=0.8\linewidth]{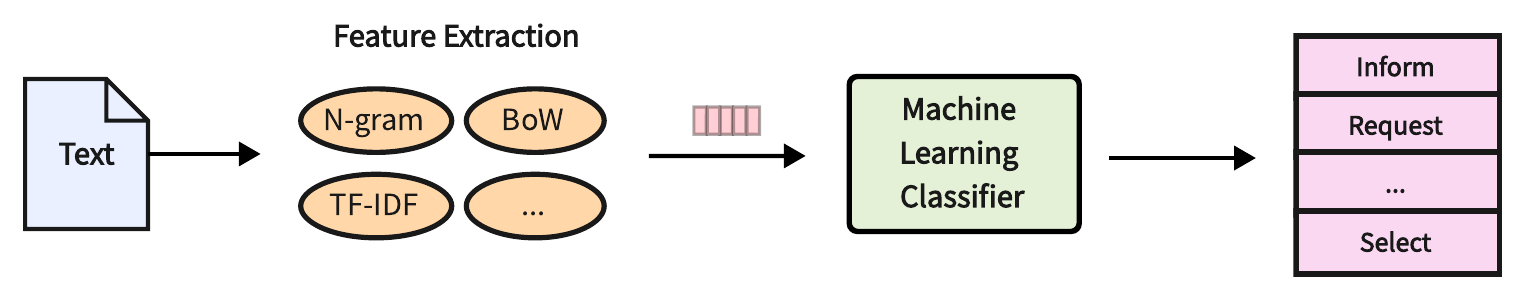} 
  \caption{Machine Learning for Text Intent Recognition.}
  \label{fig:text-machine}
\end{figure}

Early intent recognition relied on logic-based methods, which inferred intent through abductive reasoning using predefined causal rules. While interpretable, these methods lacked generalization due to their reliance on manually crafted rules. This limitation spurred the adoption of data-driven machine learning approaches, which model intent recognition as a classification problem.  As illustrated in Figure \ref{fig:text-machine}, machine learning methods leverage behavioral features to automatically infer intent, bypassing manual rule design. Supervised machine learning methods, including Naive Bayes, logistic regression, support vector machines, and decision trees, improved intent recognition by learning mappings from textual features to intent, with advancements in efficiency, generalization, and robustness~\cite{mccallum1998comparison, ng2001discriminative, genkin2007large, cortes1995support, colas2006comparison, breiman2001random, freund1996experiments}. To address challenges like limited labeled data, unsupervised and semi-supervised methods, such as clustering and graph-based learning, identified intent patterns and enabled practical applications in scenarios with sparse annotations~\cite{li2013unsupervised, padmasundari2018intent, wang2015mining, forman2015clustering}. 

The approaches of shallow machine learning techniques usually require extensive manual feature engineering and have limited performance when dealing with semantic transformations or contextual understanding. The advent of deep learning has prompted researchers to explore the application of neural network models in the domain of text intent recognition. This development has led to substantial advancements in the expressive capacity and generalization performance of these models. This section discusses the intent recognition methods based on CNN, RNN, transformer, and LLM-based network structures.

\paragraph{Traditional Neural Network Methods}

\begin{figure}[htbp]
  \centering
  \includegraphics[width=0.8\linewidth]{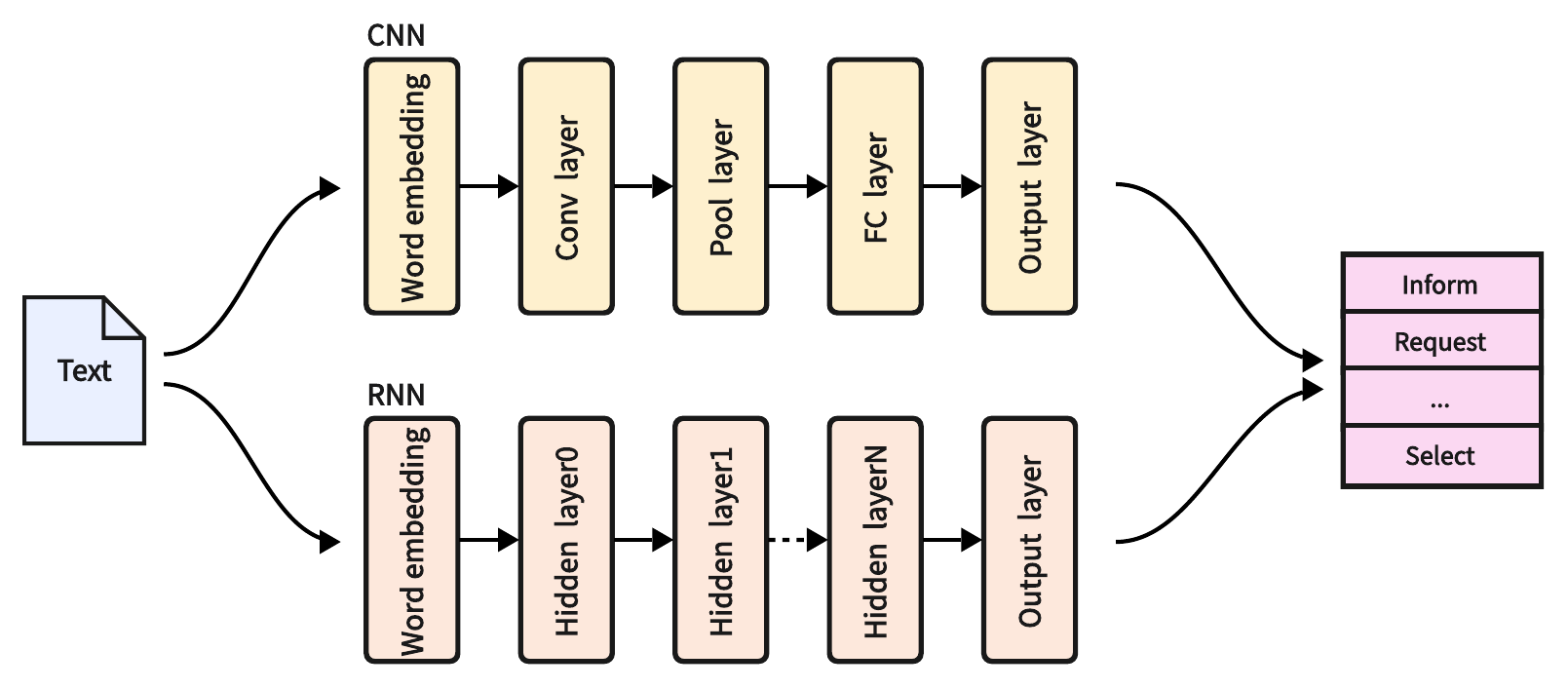} 
  \caption{The CNN/RNN-Based Model for Text Intent Recognition.}
  \label{fig:text-CNNRNN}
\end{figure}

\textbf{Convolutional neural networks (CNNs)} can effectively extract local n-gram features in sentences and exhibit excellent performance in textual intent recognition. Figure \ref{fig:text-CNNRNN} shows the CNN/RNN-based model for text intent recognition. Xu et al.~\cite{xu2013convolutional} (2013) introduced CNNs into a joint model, proposing a Triangular Conditional Random Field (Tri-CRF) model based on convolutional neural networks for simultaneous intent detection and slot filling tasks. Subsequently, Kim et al.~\cite{kim-2014-convolutional} (2014) proposed a simple CNN architecture for sentence classification, utilizing word embeddings as input and employing convolution and pooling operations to extract features for intent classification. This demonstrated the effectiveness of CNN-based feature learning in short-text intent recognition, inspiring extensive subsequent research into adopting CNN as a foundational module for intent classification.

\textbf{Recurrent neural networks (RNNs)} have attracted significant attention in the field of intent recognition due to their ability to capture sequential information. Yao et al.~\cite{yao2013recurrent} (2013) were the first to apply RNNs to language understanding tasks, processing word sequences in user utterances sequentially for intent classification. In the same year, Mesnil et al.~\cite{mesnil2013investigation} (2013) explored various RNN architectures, including bidirectional Jordan-type RNNs, for spoken language understanding, further confirming the advantages of recurrent networks in sequence modeling. However, simple RNNs suffer from limitations such as gradient vanishing. To address the issue of long-term dependencies, Hakkani-Tür et al.~\cite{hakkani2016multi} (2016) designed a bidirectional LSTM network for multi-domain joint intent and slot-filling models, enabling end-to-end semantic parsing on virtual assistant datasets and achieving notable performance gains over single-task models. Additionally, Liu and Lane\cite{liu2016attentionbased} (2016) used an attention-based encoder-decoder framework, in which the encoder uses a bidirectional RNN (BiRNN) to generate hidden states, and jointly models the intent detection and slot filling tasks by sharing the bidirectional hidden states of the encoder. Subsequently, Goo et al.~\cite{goo2018slot} (2018) introduced an attention-based slot-gated mechanism to incorporate intent context into slot recognition, proposing a slot-gated bidirectional RNN model. This model improved sentence-level semantic frame accuracy by 4.2\% and 1.9\% on the ATIS and Snips benchmark datasets, respectively, setting new state-of-the-art results at the time.

\begin{figure}[htbp]
  \centering
  \includegraphics[width=0.8\linewidth]{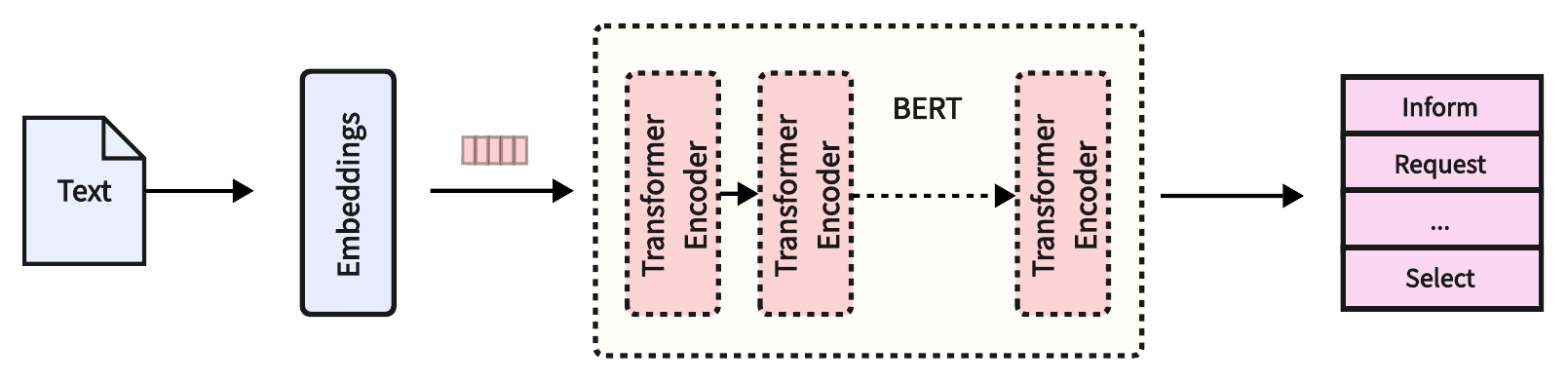} 
  \caption{The BERT Model for Text Intent Recognition.}
  \label{fig:text-bert}
\end{figure}

\textbf{Transformer model} was introduced by Vaswani et al.~\cite{vaswani2017attention} in 2017, leverages a self-attention-based encoder-decoder architecture to model long-distance dependencies in parallel, demonstrating exceptional performance in sequence modeling tasks. Since 2018, numerous Transformer-based pre-trained language models have been applied to intent recognition. A notable example is the BERT model proposed by Devlin et al.~\cite{devlin2019bert}, which employs a bidirectional Transformer encoder pre-trained on massive corpora and fine-tuned for downstream tasks. BERT has achieved unprecedented high performance across a wide range of tasks, including intent classification. Figure \ref{fig:text-bert} shows the framework of BERT for text intent recognition. Chen et al.~\cite{chen2019bert} proposed a Joint BERT model, which integrates BERT into a joint intent detection model, thereby improving intent classification accuracy to over 97\% on public datasets such as ATIS and 98.6\% on Snips, significantly surpassing the performance of traditional RNN-attention and slot-gated models. Comi et al.~\cite{comi2023zeroshotbertadapters} proposed Zero-Shot-BERT-Adapters based on the Transformer architecture fine-tuned with Adapters. This method demonstrates significant advantages in zero-shot learning scenarios, effectively improving the model's performance in classifying unknown intent across multiple languages. The success of Transformer-based models highlights that pre-trained deep self-attention representations can effectively capture intent semantics, substantially enhancing the generalization capability of dialogue intent recognition.

\paragraph{\textbf{LLM-Based Methods}}

\begin{figure}[htbp]
  \centering
  \includegraphics[width=0.8\linewidth]{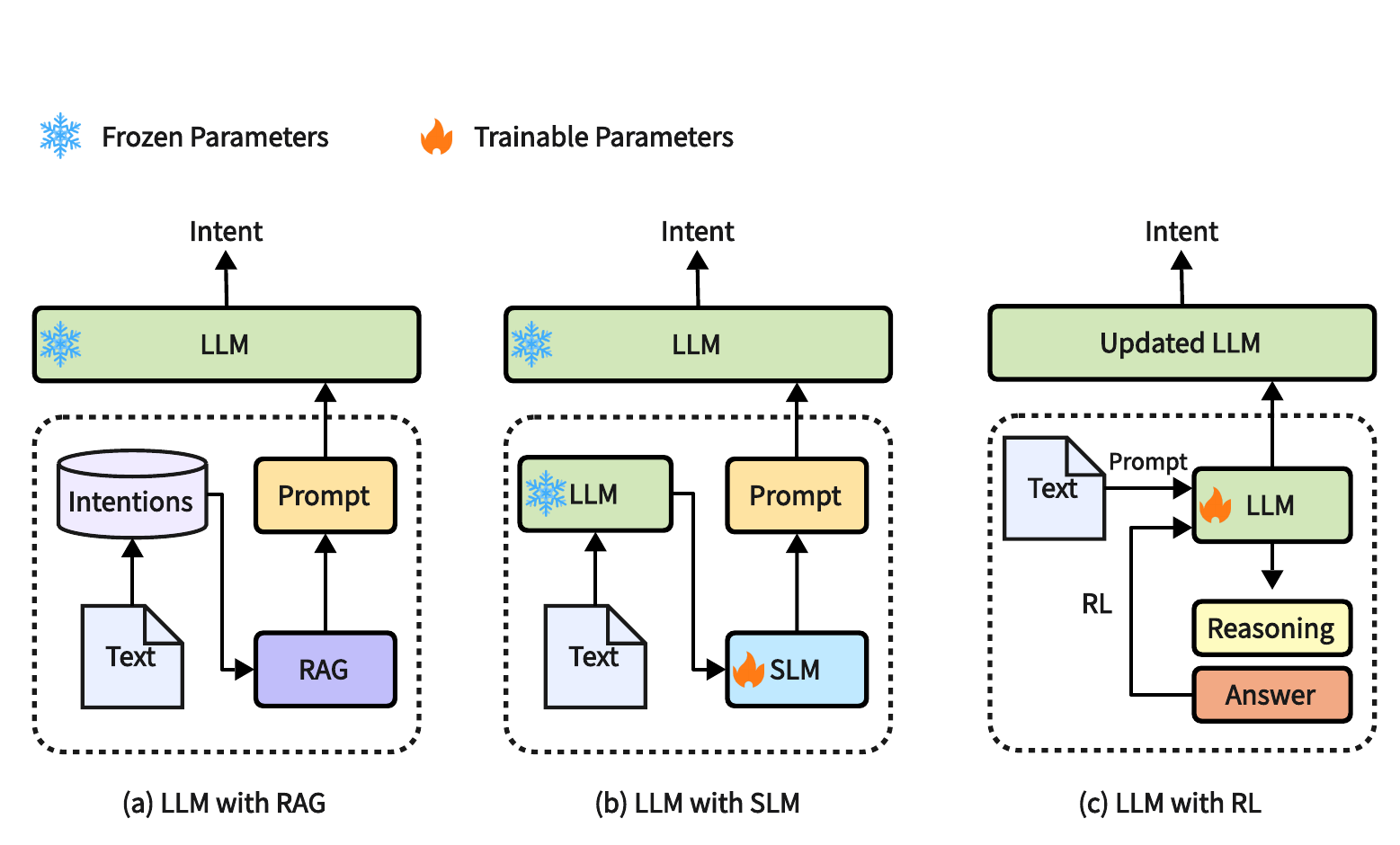} 
  \caption{The LLM-Based Models for Text Intent Recognition. (a) Leverages RAG (Retrieval-Augmented Generation) to concatenate retrieved examples with multi-turn dialogue context, constructing prompts that are fed into a frozen LLM to produce final intent classification. (b) The SLM and LLM jointly refine semantics and prompts, which are subsequently used by the frozen LLM for final classification. (c) Optimizes the intent detection capability of an LLM (as base model) through RL, yielding an updated LLM.}
  \label{fig:text-LLM}
\end{figure}

In recent years, large language models have brought breakthroughs to the field of text intent recognition with their outstanding context understanding and generation capabilities. Through innovative prompt engineering and in-context learning technologies, LLMs can significantly improve intent recognition performance in multi-turn dialogue scenarios while effectively addressing intent recognition challenges in zero-shot and few-shot learning. Figure \ref{fig:text-LLM} shows the comparison of three text intent recognition architectures based on LLM. These LLM-based approaches not only significantly enhance the model's semantic understanding depth but also markedly improve its generalization capabilities across domains and languages, providing a more flexible and efficient solution for intent recognition tasks. 

Traditional methods are constrained by predefined static intent label systems, making it difficult to distinguish between semantically similar intentions. Park et al.~\cite{park2024dynamic} proposed a dynamic label refinement approach that retrieves semantically similar dialogue examples and leverages LLMs to dynamically generate more descriptive and discriminative intent labels. This dynamic semantic decoupling strategy effectively reduces confusion among similar intentions. Liu et al.~\cite{liu2024lara} further developed the LARA framework by generating candidate intentions and then using semantic retrieval to construct a dynamic context, effectively solving the intention-context coupling problem in dialogue systems.

For small-sample learning scenarios, the IntentGPT\cite{rodriguez2024intentgpt} framework innovatively combines few-example and context-prompt generation techniques, providing semantically relevant reference labels to significantly improve the model's accuracy in recognizing unknown intent. Liang et al.~\cite{liang2024synergizing} proposed the LLM-SynCID framework for conversational intent discovery, which leverages the deep semantic understanding capabilities of large language models and the operational agility of \textbf{small language models (SLMs)} to address the issue of SLMs being unable to label newly discovered intent. 

Recently, the combination of reinforcement learning (RL) and LLM has been shown to increase the generalizability of text intent recognition. Feng et al.~\cite{feng2025improving} proposed an approach combining Reinforcement Learning with LLMs for intent detection, integrating Group Relative Policy Optimization (GRPO) with Reward-based Curriculum Sampling (RCS) while utilizing Chain-of-Thought (COT) reasoning to enhance comprehension of complex intent. This method achieved 93.25\% accuracy on the MultiWOZ dataset. For multi-turn dialogue scenarios, Liu et al.~\cite{liu2024intentaware} proposed a novel mechanism combining Hidden Markov Models (HMM) and LLM to generate context-aware, intent-driven dialogues through self-play. By introducing an auxiliary answer ranking task, the model learns more robust representations, thereby improving the accuracy of multi-turn intent classification.

\paragraph{\textbf{Discussion:}}
As a core technology of dialog systems, text intent recognition aims to accurately understand user intent through semantic analysis. From the perspective of technology development, the existing methods mainly evolve along two technical routes: machine learning and deep learning. Machine learning methods extract text features through feature engineering and construct classification models with the help of supervised learning, and at the same time combine with semi-supervised or unsupervised learning strategies to alleviate the pressure of data annotation. Although these methods perform well in specific domains, they are still limited by the complexity of feature engineering and the lack of cross-domain adaptability. Deep learning methods achieve technological breakthroughs through end-to-end representation learning, where convolutional neural networks effectively capture phrase-level semantic features through a local-awareness mechanism, recurrent neural networks and their variants excel at modeling temporal dependencies in conversations, and a pre-trained model based on the Transformer achieves, through the mechanism of self-attention, a deep contextual understanding, which significantly improves the performance of intent recognition in multi-round dialog scenarios. In recent years, LLM-based approaches have further advanced the field of intent recognition. These models acquire generalized language understanding capabilities through large-scale pre-training on massive datasets and enable zero-shot or few-shot intent recognition via prompt engineering and in-context learning.

\subsection{Vision Intent Recognition}
Visual intent recognition fundamentally differs from traditional image classification, as it requires extracting abstract semantic concepts rather than identifying concrete visual entities. Unlike object-centric taxonomies based on physical attributes or scene layouts, intent recognition operates on a higher cognitive level with weak associations to low-level visual cues, posing unique challenges for modeling latent semantics.

Early efforts in visual intent recognition focused primarily on the analysis of persuasive or motivational cues in human-centric images. The Visual Persuasion dataset proposed by Joo et al.~\cite{joo2014visual} identifies nine types of persuasive intentions, with visual cues grouped into facial expressions, body gestures, and background scenes. Building on this, subsequent studies further explored the discriminative power of facial features~\cite{joo2015automated} and scene-level attributes~\cite{huang2016inferring}, showing that different visual regions contribute complementary information for intent inference.

In a related direction, the task of action motivation aims to uncover the reasons behind human actions depicted in images. Pirsiavash et al.~\cite{pirsiavash2014inferring} was among the first to introduce this problem, emphasizing the importance of prior knowledge in understanding human behavior. To address this, Vondrick et al.~\cite{vondrick2016predicting} leveraged large-scale language models to extract commonsense priors from textual corpora, bridging the gap between visual input and abstract motivation. Furthermore, Synakowski et al.~\cite{synakowski2021adding} pioneered the integration of action intentionality into 3D vision tasks, devising an unsupervised algorithm that harnesses commonsense knowledge, including self-propelled motion (SPM), Newtonian motion, and their causal relationships, to infer whether an agent's behavior is intentional or unintentional from its 3D kinematic trajectories.

\begin{figure}[htbp]
  \centering
  \includegraphics[width=0.75\linewidth]{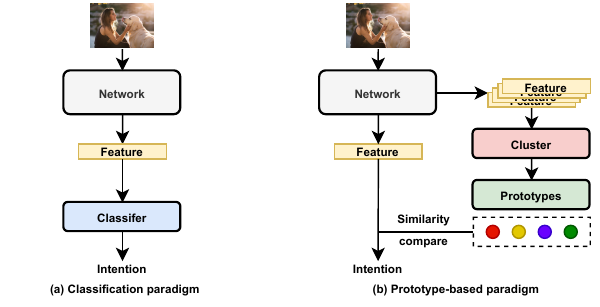} 
  \caption{The Classification Paradigm and Prototype-Based Paradigm for Vision Intent Recognition.}
  \label{fig:vision}
\end{figure}

While the aforementioned works center on human intent, recent research has expanded to include broader semantic categories beyond people, such as animals, objects, and scenes. A major milestone in this direction is the development of the Intentonomy benchmark by Jia et al.~\cite{jia2021intentonomy}, which introduces a taxonomy grounded in social psychology, comprising 28 fine-grained intent categories. This dataset redefined the problem by anchoring intent labels in psychologically meaningful constructs.
To effectively utilize such structured intent categories, Wang et al.~\cite{wang2023prototype} proposed PIP-Net, a prototype-based learning framework that constructs representative prototypes for each intent class. As illustrated in Figure \ref{fig:vision}, this approach differs from the traditional classification paradigm by explicitly modeling class representations. During training, boundary samples are filtered, and class prototypes are dynamically updated using clustering techniques. The model performs intent prediction by measuring similarity to these prototypes while enforcing inter-class separability through contrastive objectives, alleviating the label ambiguity problem prevalent in intent tasks.
Building on this foundation, Shi et al.~\cite{shi2023learnable} introduced a learnable hierarchical label embedding approach. Their method employs a multi-level Transformer to simulate the hierarchical taxonomy of intent, capturing features at coarse, intermediate, and fine levels. By leveraging coarse-grained predictions to supervise fine-grained label learning, this framework enhances discriminability and robustness in intent classification.
Further refining this line of work, Shi et al.~\cite{shi2024label} proposed the LabCR method, which disentangles multiple intentions for precise distribution calibration and employs inter-sample correlations to align instance pairs, effectively addressing label shifts and enhancing intent consistency.
To cope with the large visual diversity within intent categories, Tang et al.~\cite{tang2025MCCL} introduced the MCCL framework, which leverages multi-grained visual clues to systematically compose intent representations, achieving state-of-the-art performance on the Intentonomy and MDID~\cite{kruk2019MDID} datasets.

In summary, these efforts collectively illustrate the continuous evolution of visual intent recognition. The research trajectory has moved from early explorations focusing on human-centric scenarios, such as persuasive intent and action motivation, toward more comprehensive semantic understanding across diverse image contexts.

\subsection{Audio Intent Recognition}

Audio intent recognition (Audio-to-Intent, AIR) is typically formulated within the broader framework of Spoken Language Understanding, which aims to directly infer user intent from raw audio signals. Traditional SLU systems often employ a pipeline architecture consisting of an Automatic Speech Recognition (ASR) module followed by a NLU module. In this setup, the ASR transcribes speech into text, and the NLU module subsequently extracts intent and slot information. However, such modular systems suffer from potential mismatches in training objectives and cumulative error propagation across stages.
To overcome these limitations, recent research has increasingly focused on end-to-end SLU (E2E-SLU), which aims to directly map audio input to semantic representations. This direction is particularly motivated by audio's ability to convey both linguistic content and rich paralinguistic cues. Unlike text, audio can capture prosody, emotion, and speaker attitude, which is crucial in cases involving semantic ambiguity, emotional fluctuation, or unclear contextual boundaries.

Early efforts toward E2E-SLU include the work of Chen et al.~\cite{chen2018spoken}, who proposed a model that bypasses both ASR and textual representations by directly mapping raw audio waveforms to intent classes. Their experiments show that both training-from-scratch and fine-tuning strategies outperform conventional pipeline-based methods, validating the effectiveness of direct modeling and the reduction of error propagation.
Nevertheless, E2E models often rely on large-scale labeled data, which is costly to acquire. To address this challenge, Tian and Gorinski~\cite{tian2020improving} adopted the Reptile meta-learning algorithm to enhance generalization in low-resource settings. Their approach achieved consistent gains across four diverse datasets, demonstrating the potential of meta-learning to improve AIR under data-scarce conditions.
A key resource that has supported the advancement of AIR is the SLURP dataset, introduced by Bastianelli et al.~\cite{bastianelli2020slurp}. SLURP offers rich semantic annotations across three hierarchical levels—scenarios, actions, and entities—enabling fine-grained intent understanding and facilitating the development of more comprehensive SLU models.
With the advent of self-supervised pre-trained models, the focus has also shifted toward leveraging powerful representations. Wang et al.~\cite{wang2021fine} systematically evaluated the performance of fine-tuning Wav2vec 2.0~\cite{baevski2020wav2vec} and HuBERT~\cite{hsu2021hubert} for downstream SLU tasks. This method achieved 89.38\% accuracy on the SLURP dataset. Their results indicate significant gains through fine-tuning, but also reveal performance degradation when models are further adapted for ASR-specific tasks, suggesting a trade-off between recognition accuracy and semantic preservation.

To mitigate the limitations of audio-only inputs, several studies have proposed integrating semantic information directly into AIR systems. For example, Jiang et al.~\cite{jiang2021knowledge} presented a multi-level Transformer teacher–student distillation framework that transfers knowledge from a BERT language teacher to a speech-only Transformer student, enabling robust end-to-end intent recognition without ASR. Likewise, Higuchi et al.~\cite{higuchi2022bert} proposed BERT-CTC, utilizing BERT’s contextual embeddings as input to a CTC-based model to inject semantic context into ASR decoding. These methods demonstrate that enriching acoustic models with high-level semantics can improve both recognition and understanding performance.

Moving beyond token-level semantics, Dighe et al.~\cite{dighe2023audio} proposed a hybrid acoustic-textual model at the subword level. Their approach fuses CTC-generated subword posterior probabilities with CBOW-based semantic embeddings, striking a balance between semantic granularity and model compactness. This design enhances robustness and intent recognition accuracy, particularly in noisy environments.
Another important challenge in AIR is zero-shot intent classification. To this end, Elluru et al.~\cite{elluru2023generalized} introduced a multimodal teacher-student framework that synthesizes audio embeddings from a few example textual utterances using a neural audio generator. Combined with a pre-trained audio encoder, their model significantly improves zero-shot performance on both public and proprietary datasets.
Finally, Dong et al.~\cite{dong2025unbiased} tackled the issue of modality bias in multimodal AIR systems, where models tend to over-rely on text while underutilizing audio. They proposed MuProCL, a prototype-based contrastive learning framework that enhances audio representation via cross-audio context augmentation and semantic alignment. Experiments on SLURP and MintRec demonstrate that MuProCL not only outperforms state-of-the-art methods but also reduces modality imbalance, yielding a more interpretable and robust intent recognition model.

\begin{figure}[htbp]
  \centering
  \includegraphics[width=0.95\linewidth]{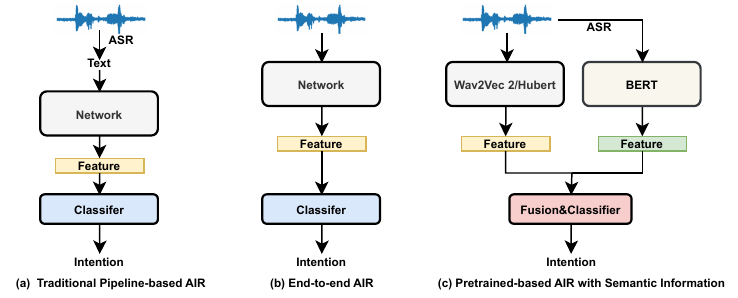} 
  \caption{The Three Frameworks for Audio Intent Recognition (AIR).}
  \label{fig:audio}
\end{figure}

In conclusion, audio intent recognition has evolved from traditional pipeline-based SLU systems to end-to-end architectures that leverage both acoustic and semantic features, as depicted in Figure \ref{fig:audio}. This progression addresses critical issues such as error propagation and semantic loss. Advances in pre-training, meta-learning, and semantic integration have significantly improved performance, especially under low-resource and noisy conditions. Additionally, emerging solutions targeting zero-shot generalization and modality imbalance underscore the importance of aligning audio representations with semantic intent. These developments collectively contribute to more accurate, robust, and efficient audio-based intent understanding.

\subsection{EEG Intent Recognition}

Electroencephalogram (EEG) technology has garnered significant attention in recent years as a non-invasive, high-temporal-resolution technique for capturing neural activity. It has proven particularly effective in identifying a person’s active intent, paving the way for various applications in brain-computer interface (BCI) and human-computer interaction (HCI) systems.

Most intent recognition studies leveraging EEG signals have primarily focused on motor (motion) intent recognition. This area has become a core challenge in EEG-based BCI systems, given its critical role in rehabilitation, prosthetics, and assistive robotics. However, motor imagery EEG (MI-EEG) signals are often noisy, exhibit high inter-subject variability, and possess complex spatial-temporal dependencies. 
Chen et al.~\cite{chen2018multitask4MIR} introduced MTLEEG, a multi-task deep recurrent neural network that concurrently processes multiple EEG frequency bands, capturing richer frequency-dependent features and significantly enhancing multi-class intent recognition. To better extract underlying spatiotemporal patterns, Zhang et al.~\cite{Zhang2018EEG-basedIR} proposed a convolutional-recurrent neural network architecture that directly processes raw EEG signals, yielding improved cross-subject generalization and practical utility in real-world BCI applications. 
Further improving generalization capabilities, Zhang et al.~\cite{zhang2019learning} developed G-CRAM, a graph-based attention network that integrates spatial, temporal, and contextual attention mechanisms. Their model consistently outperformed state-of-the-art methods on several public motor intent recognition datasets, demonstrating strong robustness to subject variability. To tackle computational inefficiencies and channel redundancy, Li et al.~\cite{Li2020GradCAM} incorporated Grad-CAM into a hybrid recurrent-convolutional architecture to perform attention-based EEG channel selection, striking a balance between classification performance and computational resource usage. 
To better exploit EEG spectral information, Idowu et al.~\cite{IDOWU2021MIR} proposed a stacked model that combines LSTM and autoencoders with t-SNE~\cite{van2008t-SNE} for feature compression, proving particularly effective in prosthetic control tasks requiring precise temporal dynamics. 
More recently, to develop cost-effective and deployable BCI systems, Li et al.~\cite{Li2024actasyouthink} proposed a BiLSTM-based PerBCI architecture compatible with low-cost EEG acquisition devices. Their system demonstrated reliable real-time performance in daily home service scenarios, especially for elderly users. To enhance modeling of the temporal-spatial-spectral structure of EEG signals, Yan et al.~\cite{yan2025temporal} introduced TSE-DA-AWS, a dual-branch architecture that performs dynamic aggregation and adaptive frequency weighting, consistently outperforming baselines on multiple public datasets and validating its effectiveness in complex motor intent recognition tasks.

Beyond motor (motion) intent recognition, some works have explored using EEG signals for other types of intent recognition. For instance, Kang et al.~\cite{KANG2015humanIR} investigated implicit intent recognition by identifying transitions in cognitive states using phase synchrony metrics like Phase Locking Value (PLV). This work demonstrated the feasibility of distinguishing between different user goals, such as navigational versus informational intent. In the domain of engineering design, Li et al.~\cite{LI2025ViT4EEGIR} applied a vision transformer (ViT)~\cite{dosovitskiy2020ViT} model, leveraging spatial-frequency EEG representations to decode design intentions, showing high versatility and accuracy in complex multitask environments.

In summary, EEG-based intent recognition has evolved from early signal-driven approaches to advanced deep learning frameworks that effectively capture the rich spatiotemporal and spectral patterns in neural activity. While unimodal approaches have laid the groundwork by establishing the feasibility of EEG for intent recognition, domain-specific models have significantly improved generalizability, robustness, and real-time applicability. Collectively, these advancements underscore the potential of EEG as a powerful modality for intent inference across both cognitive and motor domains, enabling more intelligent and responsive BCI systems.

\section{Multimodal Intent Recognition}

\begin{figure}[htbp]
  \centering
  \includegraphics[width=1\linewidth]{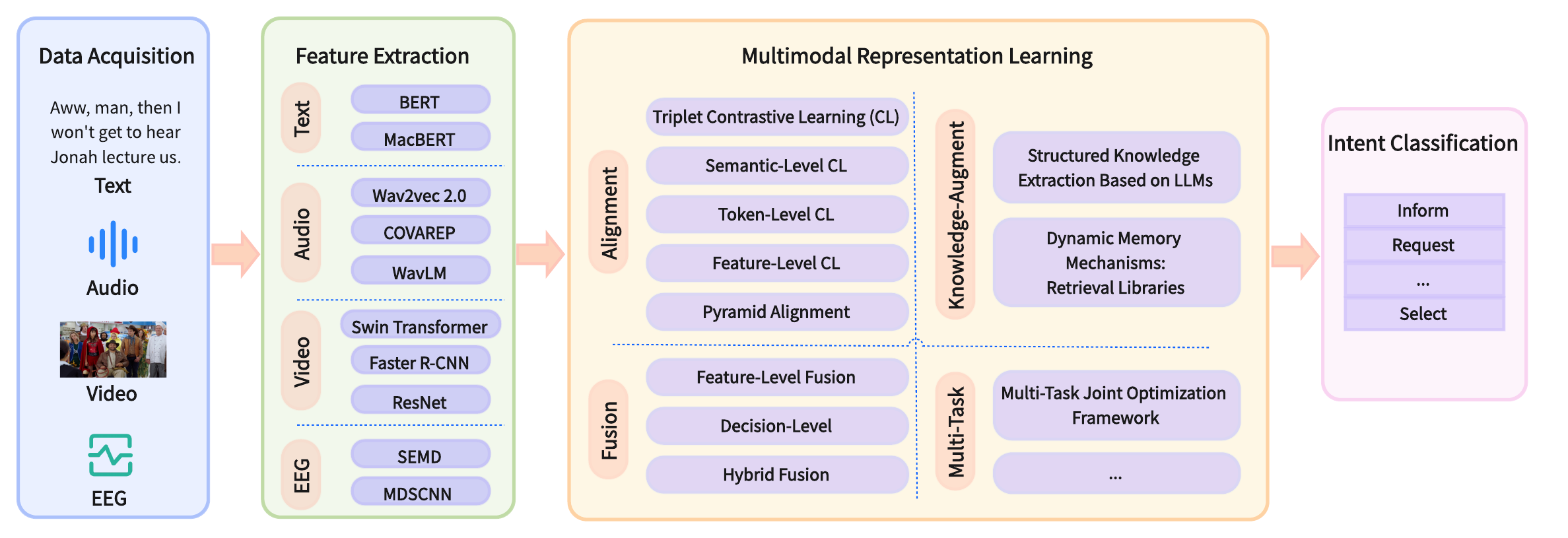}
  \caption{A Deep Learning Pipeline for Multimodal Intent Recognition.}
  \label{fig:Multimodal}
\end{figure}

With the increasing complexity of human-computer interaction scenarios, unimodal intent recognition can no longer meet the needs of real-world scenarios. Multimodal intent recognition can capture the explicit and implicit intent of the user more comprehensively by fusing multiple sources of information, such as text, audio, vision, and even physiological signals, which significantly improves the understanding accuracy and robustness. To systematically categorize research in multimodal intent recognition, we introduce a three-stage processing pipeline comprising Feature Extraction, Multimodal Representation Learning, and Intent Classification. Among these, representation learning serves as the core of multimodal modeling, where different modalities are aligned, fused, and jointly encoded to form a unified representation.
Within this stage, we further classify existing approaches into four key methodological paradigms: Fusion Methods, Alignment \& Disentanglement Methods, Knowledge-Augmented Methods, and Multi-Task Coordination Methods. This taxonomy captures the fundamental design philosophies behind current MIR systems and reflects how they address challenges such as modality interaction, semantic alignment, external knowledge incorporation, and auxiliary task coordination.
The overall pipeline is illustrated in Figure~\ref{fig:Multimodal}, and in the following subsections, we provide an in-depth overview of each category, highlighting representative methods and their key contributions.

\subsection{Fusion Methods}
\begin{figure}[htbp]
  \centering
  \includegraphics[width=1\linewidth]{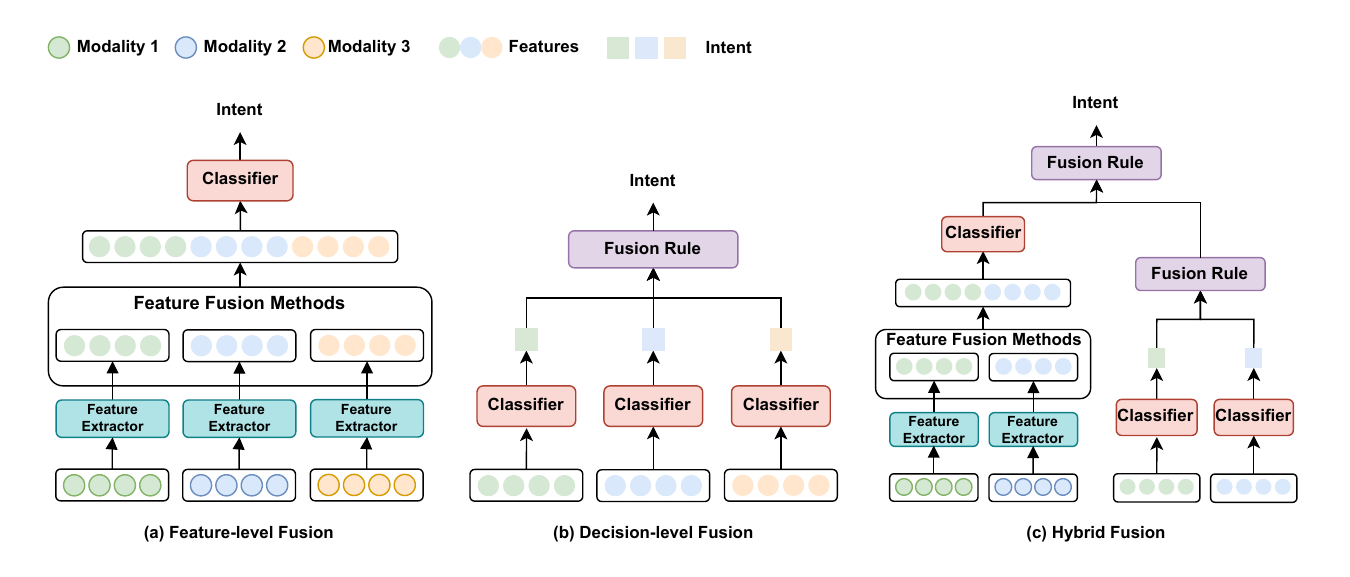} 
  \caption{The Basic Modal Fusion Methods of MIR.}
  \label{fig:fusion}
\end{figure}

The core challenge of multimodal intent recognition lies in effectively integrating complementary information from heterogeneous modalities, where the choice of fusion strategy directly determines the system's capability to infer users' latent intent. Based on the abstraction level of modal interactions, this paper categorizes existing fusion approaches into three classes: Feature-level fusion that combines raw modal features at input or intermediate layers, preserving fine-grained interactions but requiring strict alignment; Decision-level fusion that processes modalities independently before aggregating predictions; and Hybrid fusion that hierarchically combines their advantages, such as synergistic architectures integrating early feature fusion with late decision optimization. Figure \ref{fig:fusion} shows these three basic modal fusion categories.

\subsubsection{Feature-Level Fusion}
Multi-modal feature fusion methods based on attention mechanisms have demonstrated significant advantages in intent recognition tasks, effectively capturing cross-modal semantic associations, and have become a hot topic of current research. Maharana et al.~\cite{maharana-etal-2022-multimodal} introduced a cross-modal attention mechanism to interact between text feature and video feature, extracting richer feature representations for MIR. MMSAIR~\cite{shi2024impact} and MBCFNet~\cite{LI2024MBCFNET} used a multi-head attention mechanism for feature fusion, which can automatically learn the importance weights of different feature dimensions. E\textsuperscript{2}FNet~\cite{Jiang2024E2FNet} adopted a cross-attention mechanism and multiscale separable convolution to fuse EEG and EMG modalities and construct a MIR network. Hu et al. ~\cite{MVCL-DAF} proposed a dynamic attention allocation fusion (MVCL-DAF) method that dynamically allocates attention between single and multiple modalities through sample-adaptive weight allocation to adjust the integration of multimodal information. Additionally, Wang~\cite{wang2025mgc} used an adaptive mechanism-enhanced gate feature fusion module to dynamically adjust the fusion weight of each modal feature. SDIF-DA~\cite{Sdif-da}, CaVIR~\cite{IntentQA}, IntCLIP~\cite{yang2024synergy}, and EI\textsuperscript{2} network~\cite{MC-EIU} also adopted an attention mechanism for feature fusion.

Some MIR methods integrate features of different modalities through various fusion strategies. Early works ~\cite{Park2014EEG7&EM, SLANZI2017eyeEEGMIR} combined EEG and eye movement features as input to a classifier for intention prediction. Zhang et al.~\cite{zhang2020towards} proposed a feature fusion strategy based on graph neural networks to identify marketing intentions combined with images and text. In addition, ~\cite{Li2024MIREEG} effectively integrated EEG and sEMG features through TFDP and different fusion coefficients. Drawing inspiration from information bottleneck and multi-sensory processing, Zhu et al.~\cite{InMu-Net} designed InMu-Net, a multimodal intent detection framework to mitigate modal noise, redundancy, and long-tailed intent label distribution issues. The method achieved 76.05\% accuracy on the MIntRec dataset. In addition, by introducing a dynamically weighted fusion network and a multi-granular learning approach, MIntOOD~\cite{zhang2024multimodal} effectively combines text, video, and audio features, substantially advancing intent recognition and OOD detection capabilities.

\subsubsection{Decision-Level Fusion}
In decision-level fusion, opinion pooling is a core method for improving the robustness of final decisions by integrating the predicted probabilities or confidence levels of multiple classifiers or modalities. After obtaining the probability of each modality, ~\cite{trick2019multimodal} and ~\cite{zhao2024learning} used the opinion pool method to obtain the final identification. Additionally, the GIRSDF framework proposed by Yang et al. ~\cite{Yang2023GIR} extracts features using the Gaze-YOLO network and uses a sequence model for decision fusion to ultimately identify the intent of grabbing.

\subsubsection{Hybrid Fusion}
The hybrid fusion method improves the overall performance of multimodal intent recognition by integrating feature-level and decision-level fusion in a layered manner. Trick et al.~\cite{trick2023can} employed a hybrid fusion method to fuse visual and audio modalities. They trained unimodal classifiers for each body pose and speech and multimodal classifiers, either using feature fusion or decision fusion of the unimodal classifiers’ outputs with the fusion method IOP.

\subsection{Alignment \& Disentanglement Methods}
In multimodal intent recognition, modal alignment addresses the coordination of cross-modal data in temporal, spatial, and semantic dimensions, establishing a unified representation for subsequent fusion. Meanwhile, modal decoupling disentangles shared and private information from the aligned features, mitigating inter-modal interference and enhancing model interpretability. Together, these mechanisms synergistically improve the accuracy and robustness of intent recognition.

Contrastive learning, whose core idea is to bring semantically similar instances closer together in a shared embedding space while pushing unrelated samples apart \cite{hu2024comprehensive}, has become the mainstream method for aligning heterogeneous modalities~\cite{IntentQA,CAGC,TCL-MAP,A-MESS,dong2025unbiased,wang2025mgc,MVCL-DAF,gao2025EcomMIR}. In multimodal intent recognition, most methods adopt contrastive learning to reduce the semantic gap between modalities by encouraging intra-modal features to cluster tightly while ensuring inter-modal features remain distinguishable. CaVIR~\cite{IntentQA} aligns key node features in the contextual context with positive and negative samples through contrastive learning. CAGC~\cite{CAGC} utilizes global fusion context features to guide the process of contrastive learning. TCL-MAP~\cite{TCL-MAP} applies token-level contrastive learning between normal and augmented token pairs, pulling together tokens from the same pair while pushing away those from different pairs. Subsequently, A-MESS~\cite{A-MESS} synchronizes multimodal representations with label description information through triplet contrastive learning, thereby learning more effective representations. The method achieved 62.39\% accuracy on the MIntRec2.0 dataset. MuProCL~\cite{dong2025unbiased} further develops a prototype-based contrastive feature learning strategy to better enhance the alignment of the fused regional and global features. MGC~\cite{wang2025mgc} achieves dynamic alignment of multimodal data through modal mapping coupling modules.

Cross-modal attention mechanism also plays a significant role in multimodal alignment, where attention weights dynamically model semantic relationships across modalities, enabling explicit fine-grained interaction modeling \cite{Sdif-da,ye2023cross,MC-EIU}. SDIF-DA \cite{Sdif-da} resolves the issue of multimodal feature heterogeneity by employing cross-modal attention mechanisms in its shallow interaction module to respectively align video and audio features with corresponding text features. Additionally, cross-modal pyramid alignment employs a hierarchical attention mechanism to progressively align features at multiple granularities, achieving finer-grained semantic consistency across modalities. CPAD \cite{ye2023cross} aligns features from different modalities at the same level through cross-modal pyramid alignment, taking into account all corresponding features at all levels, thereby enhancing the understanding of visual intent.

To address the issues of entangled multimodal semantics with modality structures and insufficient learning of causal effects of semantic and modality-specific information, Chen et al.~\cite{DuoDN} proposed a Dual-oriented Disentangled Network (DuoDN) for MIR, which employs a dual-oriented encoder to separate representations into semantics- and modality-oriented components. LVAMoE~\cite{LVAMoE} uses a dense encoder to map different modalities onto a shared subspace, achieving explicit modal alignment. Through a sparse MoE module, LVAMoE learns features unique to each modality and uses orthogonal constraints to ensure their independence from invariant representations, thereby achieving modal decoupling.

\subsection{Knowledge-Augmented Methods}
Multimodal intent recognition faces dual challenges of semantic ambiguity and data sparsity. Knowledge-enhanced methods address these by integrating large language models for structured knowledge extraction and retrieval libraries as dynamic memory mechanisms, significantly improving contextual reasoning and few-shot generalization capabilities.

\paragraph{LLM-Based Approaches} CaVIR~\cite{IntentQA} uses the large language model GPT during the testing phase to enhance the model's common sense reasoning capabilities, thereby more accurately understanding the intent in the video. A-MESS~\cite{A-MESS} leverages the powerful language generation capabilities of LLM to generate three different interpretations or descriptions for each label. It then uses the label descriptions generated by LLM as positive samples and the descriptions of other labels as negative samples to construct a triplet contrast learning model, thereby enhancing the model's semantic understanding capabilities. Zhang et al.~\cite{zhang2024content} used LLM to solve ambiguity issues in image intent perception tasks by generating visual prompts to guide LLMs to better understand the intent information in images. ~\cite{gao2025EcomMIR}, ~\cite{song2025MIRdialogue}, and ~\cite{li2025CuSMer} also use multimodal LLMs to deeply integrate text and visual information and accurately identify user intent. In addition, KDSL~\cite{chen2025KDSL} effectively solves the problem of inter-task performance constraints in few-shot multimodal intent recognition tasks by using smaller LLMs to convert knowledge into interpretable rules and combining them with larger LLMs for collaborative prediction.

\paragraph{Retrieval-Based Approaches} Within the Knowledge-Augmented Methods category, retrieval-augmented approaches in large language models enhance intent understanding by retrieving relevant textual knowledge, while retrieval-based multimodal intent recognition methods leverage similar mechanisms to fetch contextual information from modalities like video and audio, thereby enriching feature representations for improved performance. The Cross-video Bank in CAGC~\cite{CAGC} is mainly used to store scenes that are highly similar to the current video and provide more accurate cross-video contextual information, thereby enhancing intent understanding. MuProCL~\cite{dong2025unbiased} utilized external context information (Top-k samples) from the training set to improve model performance. By introducing context information from similar audio samples, we enhanced the representational power of audio features.

\subsection{Multi-Task Coordination Methods}
\begin{figure}[htbp]
  \centering
  \includegraphics[width=0.9\linewidth]{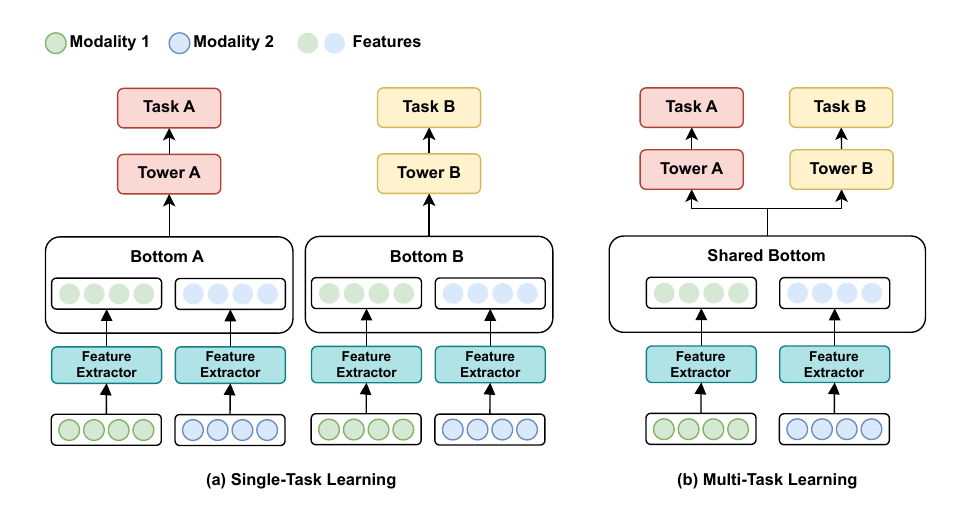} 
  \caption{Single-Task Learning and Multi-Task Learning in MIR.}
  \label{fig:multi-task}
\end{figure}

Multi-Task Coordination Methods advance multimodal intent recognition by jointly optimizing multiple related tasks within a unified framework, enabling the model to capture shared representations across tasks such as intent classification and emotion recognition. By coordinating these objectives, these methods address the limitations of single-task learning, enhancing generalization and robustness in multimodal settings. The integration of audio, text, and other modalities in a multi-task framework allows for a more comprehensive understanding of user intent, leveraging complementary information to improve overall performance. The common learning paradigms in MIR are illustrated in Figure \ref{fig:multi-task}.

EI\textsuperscript{2} network~\cite{MC-EIU} jointly models sentiment and intent recognition through a multi-task learning framework. The model utilizes multimodal inputs (text, audio, video) and dialogue history information, capturing the deep connections between sentiment and intent through task-specific encoders and interactive attention mechanisms. In addition, the publicly available MC-EIU dataset fills the data gap in multimodal, multilingual sentiment-intent joint understanding, providing an important resource for related research. MMSAIR~\cite{shi2024impact} also employs a multi-task learning framework to jointly model sentiment analysis and intent recognition. It leverages multimodal inputs with shared feature encoding, followed by fusion via multi-head attention mechanisms to generate sentiment and intent representations. The model uses a weighted loss function for joint optimization, enabling mutual reinforcement between tasks. In addition, MBCFNet~\cite{LI2024MBCFNET} used a multi-task learning framework to jointly optimize emotion recognition as an auxiliary task with intent recognition, utilizing multimodal data from EEG, audio, and text to significantly improve the accuracy of intent recognition. Experiments demonstrated that the introduction of emotional information and EEG signals can effectively solve the problem of intent ambiguity in the same text in different contexts, providing new ideas for multimodal human-computer interaction.

\begin{table}[htbp]
  \centering
  \caption{Summary of Representative Approaches of Multimodal Intent Recognition}
  \label{tab:MIR-multimodal}
    \begin{tabular}{p{2.2cm} p{1.6cm} c p{5cm} p{2.2cm}}
    \toprule
    \textbf{References} & \textbf{Dataset} & \textbf{Modality} & \multicolumn{1}{c}{\textbf{Method}} & \textbf{Performance} \\
    \midrule
    Ye et al.~\cite{ye2023cross} & Intentonomy$^\ddagger$ & VT & Hierarchical cross-modal alignment pyramid with label texts. & 36.50\%Ave F1 \\ 
    Yang et al.~\cite{yang2024synergy} & Intentonomy$^\ddagger$ & VT & CLIP-based LLM filtering for label text representations. & 42.66\%mAP \\ 
    Li et al.~\cite{IntentQA} & IntentQA & VT & Context-aware Video Intent Reasoning (CaVIR) model. & 57.64\% Accuracy \\ 
    Zhou et al.~\cite{TCL-MAP} & MIntRec & VAT & Token-level contrastive learning with modality-aware prompting. & 73.62\% Accuracy \\ 
    Sun et al.~\cite{CAGC} & MIntRec & VAT & Global contextual information integration across videos. & 73.39\% Accuracy \\ 
    Li et al.~\cite{LVAMoE} & MIntRec & VAT & One-tower architecture addressing cross-modal heterogeneity. & 73.13\% Accuracy \\ 
    Shen et al.~\cite{A-MESS} & MIntRec, MIntRec2.0 & VAT & Anchor-based embedding with semantic synchronization. & 74.12\% Accuracy, 62.39\% Accuracy  \\ 
    Zhu et al.~\cite{InMu-Net} & MIntRec & VAT & Information bottleneck for modality redundancy, multi-sensory processing. & 76.05\% Accuracy \\ 
    Huang et al.~\cite{Sdif-da} & MIntRec & VAT & Shallow-to-deep interaction with data augmentation. & 73.71\% Accuracy \\ 
    Hu et al.~\cite{MVCL-DAF} & MIntRec, MIntRec2.0 & VAT & Dynamic adaptation to diverse multimodal samples. & 74.72\% Accuracy, 57.80\% Accuracy \\ 
    Chen et al.~\cite{DuoDN} & MIntRec, MIntRec2.0 & VAT & Dual-oriented disentangled network with counterfactual intervention. & 75.28\% Accuracy, 57.76\% Accuracy \\ 
    Wang et al.~\cite{wang2025mgca} & MIntRec & VAT & Modal mapping coupled to cross-modal attention, dynamic alignment with gated fusion weighting. & 73.93\% Accuracy \\ 
    Zhao et al.~\cite{zhao2024learning} & Kitchen robot & V, Gaze, Gestures & Bayesian multimodal fusion with batch confidence learning. & 72.0\% Accuracy \\ 
    Li et al.~\cite{CMSLIU} & CMSLIU & ATE & Brain-computer fusion for identical-text intention disambiguation. & 69.4\% Accuracy \\ 
    \bottomrule
  \end{tabular}
  \raggedright
  $^\ddagger$ Intentonomy is originally a visual-only (V) dataset; text modality was constructed in each work respectively.
\end{table}

\subsection{Discussion}

Multimodal intent recognition has seen significant advancements, driven by large-scale datasets and deep learning architectures, with various approaches summarized in Table \ref{tab:MIR-multimodal}. Despite these developments, key challenges persist. While fusion methods effectively combine text, audio, and visual signals, temporal asynchrony and inter-modal heterogeneity complicate alignment, particularly in multi-round conversations where deep audio-visual-text fusion proves difficult. The field also grapples with open-world challenges, including out-of-scope detection and semantic ambiguities in intent mapping, which are further exacerbated by the scarcity of labeled data. Although knowledge-augmented and LLM-based approaches show promise in addressing these issues, limitations in scalability and fine-tuning persist. Future progress hinges on developing adaptive fusion techniques that overcome modality disparities while improving generalization across diverse, real-world scenarios.

\section{Performance Evaluation Metrics}
Given the evaluation requirements of intent recognition tasks, researchers typically use four core metrics to evaluate systems: \textbf{Accuracy (ACC), Precision (P), Recall (R), and F1 score (F1)}. Accuracy is defined as the proportion of correct predictions made by the model in its entirety. This metric is particularly well-suited to scenarios where the distribution of categories is balanced. Precision, on the other hand, is the proportion of samples predicted to be positive that are in fact positive, thereby reflecting the model's prediction accuracy. Recall, meanwhile, is the proportion of samples that are in fact positive and are correctly identified, thus emphasising the model's coverage. Finally, the F1 score is the harmonic average of Precision and Recall, making it suitable for tasks with an imbalanced distribution of categories. This metric can be used to evaluate the accuracy of the model. The distribution imbalance tasks facilitate a more balanced evaluation of the model's overall performance. Zhang et al.~\cite{zhang2022contrastive} utilized accuracy as the evaluation metric to validate the effectiveness of their proposed method. Wong et al.~\cite{wong2023vision} employed precision, recall, and F1-score to evaluate the intent recognition performance of their model. Likewise, Nguyen et al.~\cite{TECO} adopted accuracy, F1-score, precision, and recall as systematic evaluation metrics, effectively assessing their model's performance.

In addition, intent recognition tasks also use specialized evaluation measures like \textbf{Weighted F1-score (WF1), Weighted Precision (WP), Pearson Correlation (Corr), In-scope Classes F1-score (F1-IS), Out-of-scope Class F1-score (F1-OOS), and equal error rate (EER)}. These metrics help gauge how well a model identifies and classifies user intentions, ensuring both accuracy and robustness in real-world applications. In the study on multimodal intent recognition, Zhang et al.~\cite{mintrec2} employed a comprehensive set of evaluation metrics for in-scope intent classification, including F1-score, P, R, ACC, Weighted F1-score (WF1), and Weighted Precision (WP), to thoroughly assess the model’s performance on known intent categories. For out-of-scope intent detection, they adopted metrics commonly used in open intent recognition, such as Accuracy, Macro F1-score across all classes, F1-IS, and F1-OOS, to effectively evaluate the model’s robustness and accuracy in identifying unknown intent. The definitions for them are as follows:

\begin{equation}
Macro F1 = \frac{1}{|C|} \sum_{i \in C} \frac{2 \cdot P_i \cdot R_i}{P_i + R_i}, \quad WF1 = \sum_{i \in C} w_i \cdot \left( \frac{2 \cdot P_i \cdot R_i}{P_i + R_i} \right)
\end{equation}

\noindent where \( |C| \) is the total number of classes, \( P_i \) and \( R_i \) are the precision and recall for class \( i \), \( w_i \) is the weight for class \( i \).

\begin{equation}
F1\text{-}IS = \frac{1}{|C_{IS}|} \sum_{{is} \in C_{IS}} \frac{2 \cdot P_{is} \cdot R_{is}}{P_{is} + R_{is}}, \quad F1\text{-}OOS = \frac{2 \cdot P_{OOS} \cdot R_{OOS}}{P_{OOS} + R_{OOS}}
\end{equation}

\noindent where \( |C_{IS}| \) is the total number of in-scope classes, \(P\) denotes Precision and \(R\) represents Recall.

Pinhanez et al.~\cite{pinhanez2021using} utilized EER to evaluate out-of-scope intent detection and it measures the model's ability to distinguish between known and unknown intent. The EER is a metric used to quantify the classification error rate when the False Acceptance Rate (FAR) and the False Rejection Rate (FRR) are equal or closest. The FAR is defined as the proportion of out-of-scope intent incorrectly classified as in-scope, while the FRR is the proportion of in-scope intent incorrectly classified as out-of-scope. These two metrics are defined as:

\begin{equation}
FAR = \frac{N_{\text{acc}}^{\text{OOS}}}{N^{\text{OOS}}}, \quad FRR = \frac{N_{\text{rej}}^{\text{IS}}}{N^{\text{IS}}}
\end{equation}
 
\noindent where $N_{\text{acc}}^{\text{OOS}}$ and $N_{\text{rej}}^{\text{IS}}$ denote the number of OOS samples incorrectly accepted and IS samples incorrectly rejected, respectively, and $N^{\text{OOS}}$ and $N^{\text{IS}}$ are the total number of OOS and IS samples.

\section{Applications}

Intent recognition technology has become pervasive in a variety of practical scenarios, ranging from the conventional interpretation of user intent in natural language processing to the sophisticated reasoning that integrates multiple sources of information, such as speech, image, and behaviour, in multimodal interaction systems. Divergent application scenarios give rise to marked discrepancies in terms of the requirements and methodologies employed for intent recognition. To illustrate this point, we may consider the case of educational scenarios, wherein the emphasis is placed on the comprehension of learning motivation and the intent behind questioning. In contrast, the medical field accords greater significance to the recognition of patients' needs and the identification of rehabilitation behavioural intent. This section explores various applications of intent recognition, as illustrated in Table \ref{tab:application}.

\begin{table}[htbp]
  \centering
  \caption{Summary of Intent Recognition Applications}
  \label{tab:application}
  \begin{tabular}{l p{0.72\textwidth}}
    \toprule
    \textbf{Domain} & \textbf{Typical Applications} \\
    \midrule
    Human–computer interaction & Gesture/speech control, collaborative task intent, natural interaction \\
    Education & Motivation identification, questioning/confusing intentions, individualized tutoring \\
    Healthcare & Disease expression intent, rehabilitation action intent, mental state recognition \\
    Smart home & Appliance control intent, voice command recognition, behavioral trigger intent \\ 
    Automotive systems & Driver intent prediction, pedestrian/non-motorized behavior recognition \\
    Marketing and advertising & Purchase intent recognition, interest prediction, ad personalized recommendation \\
    Sports & Action intent prediction, tactical intent recognition, postural understanding \\
    \bottomrule
  \end{tabular}
\end{table}

\begin{enumerate}[label=(\arabic*)]

\item \textbf{Human–computer interaction}:
In human-computer interaction (HCI) scenarios~\cite{he2025conversational,jain2018recursive,lemasurier2021methods,losey2018review,pascher2023communicate}, intent recognition is primarily employed to decipher the control commands or interaction requirements articulated by human users through voice, gestures, and movements. For instance, the study~\cite{jain2018recursive} proposed a recursive method based on Bayesian filtering to recognize the user's intention and reach the purpose of collaborative task completion between robots and humans. The intent of the collaborative task is perceived through gestures or body postures, thereby enhancing the robot's capacity to discern and respond to human intent, thus facilitating natural and efficient collaborative operations.

\item \textbf{Education}:
In the domain of education, intent recognition has become a prevalent technique in intelligent tutoring systems and personalized learning platforms~\cite{yusuf2025pedagogical,pearce2023build,sinval2024correlates,otache2024entrepreneurship,pearce2023build,nguyen2024unveiling}. This method is employed to identify students' learning objectives, questioning intentions, confusion states, and learning motivations. The study~\cite{pearce2023build} categorizes the questions posed by students into predefined intent categories. The model then selects the corresponding text content as the context to answer the questions based on the recognized intent categories.

\item \textbf{Healthcare}:
Intent recognition in healthcare scenarios encompasses a multitude of dimensions~\cite{liu2017intent,zhang2013source,xu2021noninvasive,xu2024insensor,noor2024detecting,zhang2025multimodal}, including patient disease representation, rehabilitation movement control, and psychological state recognition. In the study of~\cite{zhang2013source}, the researchers analyzed the usefulness of different data sources, such as EMG signals and prosthetic kinematic parameters. By selecting the most effective combination of these data sources, they achieved real-time recognition of an amputee's movement intention, enabling autonomous prosthetic limb control.

\item \textbf{Smart home}:
The application of intent recognition technology in the smart home~\cite{rafferty2017activity,desot2019towards,tang2024temporal} allows the system to recognize the user's control needs or habitual behaviors in the current environment and automatically adjust the state of the device, thus realizing a smarter and more convenient seamless interaction experience. The study~\cite{desot2019towards} applies end-to-end (E2E) audio intent recognition technology to smart home scenarios, which can solve the limitations of traditional control methods, improve interaction convenience and user experience, and make the smart home more intelligent and humanized.

\item \textbf{Automotive systems}:
In the domain of automotive systems, the concept of intent recognition finds application in the analysis of potential behaviors and the discernment of decision-making intentions among various road users such as drivers, pedestrians, cyclists, and vehicles\cite{quintero2017pedestrian,berndt2008continuous,sharma2025predicting,li2024behavior,varytimidis2018action,fang2024behavioral,Liu2025driver_IR}. The implementation of intent recognition technologies contributes to the development of intelligent automotive systems that are designed to enhance safety and efficiency. Intent recognition is applied in advanced driver assistance systems (ADAS) in this study~\cite{berndt2008continuous}, aiming to understand driver behavior earlier and more accurately, thereby enhancing safety outcomes. The study~\cite{sharma2025predicting} improves traffic safety and reduces the risk of traffic accidents by predicting pedestrians' intention to cross the road.

\item \textbf{Marketing and advertising}:
In the domain of marketing and advertising~\cite{schultz2025consumer,mittal2024determining,zhang2021multimodal,soleymani2017multimodal,wu2025seeing,zhang2022multimodal}, intent recognition is employed to predict a user's purchase intention, points of interest, or clicking behaviors. This, in turn, enables the recommendation of personalized content and precise placement. This study~\cite{zhang2021multimodal} presented a multimodal marketing intent analysis system called MMIA that can identify and parse marketing intent by analyzing multimodal content (text and images) on social media platforms. Another study~\cite{soleymani2017multimodal} analyzed features such as facial expressions, eye movements, queries, and implicit user interactions of users with different search intentions to automatically identify users' search intent in the early stages of an image search session.

\item \textbf{Sports}:
In the context of sports training and competition, the recognition of intent is employed to facilitate comprehension of a player's tactical behaviors, technical movement, or synergistic intent~\cite{hassan2024robotics,wang2024long,li2024omniactions}. By analyzing a player's facial expression and eye movements, the system proposed in study~\cite{hassan2024robotics} can determine whether a player has foreseen and planned to act by touching the ball. This can be used to determine whether a handball incident in a soccer match is an intentional act or not.

\end{enumerate}

\section{Challenges and Future Directions}

Despite substantial progress in intent recognition, several critical challenges remain unresolved. The inherently diverse, ambiguous, and dynamic nature of human intent, along with the complexity of multi-intent utterances and the evolving intent throughout dialogues, continues to pose significant obstacles. These challenges underscore the need for the development of more advanced intent recognition systems and the continuous refinement of existing approaches. The key challenges and potential future research directions are summarized as follows:

\begin{enumerate}[label=(\arabic*)]

\item \textbf{Intent ambiguity and semantic vagueness}:
User utterances are often context-dependent, ambiguous, or semantically underspecified, which makes accurate intent inference particularly difficult. For instance, a statement such as “It’s cold in here” could imply a request to adjust the temperature, a complaint, or merely an observation, depending on factors such as user intent history, prosody, and environmental context. Addressing such ambiguity necessitates integrating contextual reasoning, pragmatic inference, and affective state modeling into intent recognition frameworks.

\item \textbf{Multi-intent and compositional intent recognition}:
Natural language allows users to express multiple or compositional intent within a single utterance. For example, the sentence “Book a table for two and remind me to call mom” conveys two distinct intentions: making a reservation and setting a reminder. Moreover, hierarchical or conditional intent structures further challenge traditional single-label classification paradigms. Addressing this requires support for multi-label classification, intent segmentation, and compositional reasoning~\cite{wu2021label,cheng2024towards,qin2025DSCP}. For instance, Qin et al.~\cite{qin2025DSCP} proposed a Divide-Solve-Combine Prompting (DSCP) strategy for multi-intent detection using large language models. Cheng et al.~\cite{cheng2024towards} utilized hierarchical attention to divide the scopes of each intent and applied optimal transport to achieve the mutual guidance between slot and intent.

\item \textbf{Intent evolution in multi-turn dialogues}:
In real-world interactions, user intentions are often dynamic and evolve across multiple dialogue turns. An initial intent such as “search for flights” may evolve into “book the cheapest option” or even “cancel everything” depending on updated constraints or emotional changes. Capturing such temporal evolution requires advanced mechanisms such as context-aware dialogue state tracking, memory-augmented models, and intent revision detection. Recent studies have begun exploring these challenges using the capabilities of large language models~\cite{abro2019multi, liu2024intent, liu2024lara}.

\item \textbf{Modal heterogeneity and asynchrony}:
Multimodal intent recognition involves heterogeneous input signals such as speech, facial expressions, gestures, text, or physiological data, each with distinct temporal resolutions and semantic representations. For instance, gaze or gesture cues may temporally precede or follow verbal expressions, while EEG signals operate on a millisecond scale. These temporal and representational disparities pose significant challenges to synchronous fusion. Addressing this requires sophisticated temporal alignment, modality-specific encoders, and attention-based fusion strategies that preserve the semantic integrity of each modality.

\item \textbf{Out-of-domain intent detection}:
In open-world scenarios, users frequently express intent outside the predefined set observed during training. Relying solely on closed-set assumptions can lead to erroneous or inappropriate system responses. Robust out-of-domain (OOD) detection is essential to maintain system reliability and user trust. Recent efforts have explored open-set recognition~\cite{zheng2020out, jin2022towards} and hybrid supervised–unsupervised approaches~\cite{zhan2021out, zhou2022knn} to distinguish unseen intent and appropriately defer, reject, or redirect such cases.

\item \textbf{Long-tail intent distribution}:
Many intent categories, particularly user-specific or domain-specific ones, are severely underrepresented in available training data. For instance, intentions like “book appointment” may have abundant instances, while rarer ones such as “report vaccine side effect” may be sparsely labeled. This long-tail distribution hinders model generalization. Recent work leverages few-shot learning, transfer learning, and synthetic data generation to address these challenges~\cite{zhang2024towards, zhou2024intentional}.


\item \textbf{Cross-lingual and cross-cultural generalization}:
The majority of intent recognition systems are trained on English-centric datasets, limiting their applicability in multilingual and multicultural settings. For instance, the intent behind the phrase “Can you recommend something spicy?” might vary across cultures. Cross-lingual transfer learning using multilingual models (e.g., XLM-R~\cite{conneau2020unsupervised}, mT5~\cite{xue2021mt5}) and the development of culturally-aware benchmarks and modeling techniques are crucial to enhance inclusiveness and global applicability~\cite{faria2025uddessho,yu2025injongo}.

\item \textbf{Continuous intent reasoning for dynamic environments}
Traditional intent recognition methods typically rely on static environment assumptions, making them inadequate for real-time changing interaction scenarios. When physical states or task requirements alter, systems must continuously update intent reasoning, placing higher demands on models' perceptual fusion and temporal processing capabilities. This challenge is particularly prominent in Embodied AI applications. For instance, service robots need to adjust operational intent based on moving objects, and autonomous driving systems require real-time updates to navigation strategies in response to sudden road conditions. Addressing these issues necessitates the development of dynamic intent reasoning mechanisms with environmental awareness and adaptive capabilities.


\end{enumerate}

\section{Conclusion}
This article systematically elucidates the evolution of intent recognition from single-modality to multimodal approaches, thoroughly examining high-performance methods based on deep learning. It covers both single-modality techniques (text, audio, visual, EEG) and advanced multimodal approaches, including feature fusion, alignment \&disentanglement, knowledge augmentation, and multi-task coordination. The article comprehensively reviews public datasets and standardized evaluation metrics in the field of multimodal intent recognition, providing researchers with a robust benchmark framework and reference standard. Intent recognition has been demonstrated to have broad application potential across diverse domains, including human-computer interaction, automotive systems, education, and healthcare. Looking ahead, with the continuous advancement of multi-source data acquisition technologies, research into key issues such as data heterogeneity, modal synchronization, and cross-domain intent detection will accelerate. These breakthroughs will significantly enhance the accuracy, robustness, and universality of multi-modal intent recognition, driving intelligent interaction systems toward smarter and more human-centric directions.



\bibliographystyle{ACM-Reference-Format}
\bibliography{main}


\end{document}